\documentclass{article}

\usepackage{microtype}
\usepackage{graphicx}
\usepackage{subfigure}
\usepackage{booktabs} 

\usepackage[breaklinks=true, hidelinks]{hyperref}

\usepackage[preprint]{icml2026}

\usepackage{amsmath}
\usepackage{amssymb}
\usepackage{mathtools}
\usepackage{amsthm}

\usepackage[capitalize,noabbrev]{cleveref}

\usepackage{multirow}
\usepackage[table]{xcolor}
\usepackage{enumitem}
\usepackage{dsfont}
\usepackage{comment}
\usepackage{algorithm}
\usepackage{algorithmic}



\usepackage{tcolorbox}
\usepackage{tikz}
\usetikzlibrary{positioning, arrows.meta, shapes.geometric}
\usepackage{makecell}

\theoremstyle{plain}

\theoremstyle{definition}

\theoremstyle{remark}

\icmltitlerunning{FedPDPO: Federated Personalized Direct Preference Optimization}

\begin{document}

\twocolumn[
\icmltitle{FedPDPO: Federated Personalized Direct Preference Optimization for \\
Large Language Model Alignment}



\icmlsetsymbol{equal}{*}

\begin{icmlauthorlist}
\icmlauthor{Kewen Zhu}{inst1}
\icmlauthor{Liping Yi}{inst1}
\icmlauthor{Zhiming Zhao}{inst1}
\icmlauthor{Zhuang Qi}{inst2}
\icmlauthor{Han Yu}{inst2}
\icmlauthor{Qinghua Hu}{inst1}
\end{icmlauthorlist}

\icmlaffiliation{inst1}{College of Intelligence and Computing,
    Tianjin University, Tianjin, China}
\icmlaffiliation{inst2}{College of Computing and Data Science,
    Nanyang Technological University, Singapore}

\icmlcorrespondingauthor{Liping Yi}{lipingyi@tju.edu.cn}

\icmlkeywords{Personalized federated learning, Direct Preference Optimization, LoRA, Large Language Models}

\vskip 0.3in
]



\printAffiliationsAndNotice{} 

\begin{abstract}
Aligning large language models (LLMs) with human preferences in federated learning (FL) is challenging due to decentralized, privacy-sensitive, and highly non–independent and identically distributed (non-IID) preference data. 
Direct Preference Optimization (DPO) offers an efficient alternative to reinforcement learning with human feedback (RLHF), but its direct application in FL suffers from severe performance degradation under non-IID data and limited generalization of implicit rewards. 
To bridge this gap, we propose \texttt{FedPDPO} (\underline{Fed}erated \underline{P}ersonalized \underline{D}irect \underline{P}reference \underline{O}ptimization), a personalized federated framework for preference alignment of LLMs. 
It adopts a parameter-efficient fine-tuning architecture where each client maintains a frozen pretrained LLM backbone augmented with a Low-Rank Adaptation (LoRA) adapter, enabling communication-efficient aggregation. 
To address non-IID heterogeneity, we devise (1) the globally shared LoRA adapter with the personalized client-specific LLM head. Moreover, we introduce (2) a personalized DPO training strategy with a client-specific explicit reward head to complement implicit rewards and further alleviate non-IID heterogeneity, and (3) a bottleneck adapter to balance global and local features. We provide theoretical analysis establishing the probabilistic foundation and soundness. Extensive experiments on multiple preference datasets demonstrate its state-of-the-art performance, achieving up to a $4.80\%$ average accuracy improvements in federated intra-domain and cross-domain settings. The code will be made publicly available upon acceptance.

\end{abstract}

\section{Introduction}
\label{submission}

Aligning large language models (LLMs) with human preferences is essential for their reliable deployment across a broad range of applications, spanning both open domains (\emph{e.g.}, dialogue, summarization, and instruction following) 
\citep{stiennon2020learning,bai2022constitutional,achiam2023gpt} and vertical domains (\emph{e.g.}, healthcare, law, and finance) \citep{singhal2023large,kasneci2023chatgpt}.

However, in practical vertical domains, human preference data are naturally distributed across decentralized users, institutions, or enterprises. 
Such data are typically prohibited from being shared due to privacy constraints, \emph{e.g.}, patient records stored by medical institutions \citep{rieke2020future} and user preference data collected by commercial enterprises \citep{kairouz2021advances}.
Federated learning (FL) \citep{mcmahan2017communication,li2020federated2}, a privacy-preserving distributed learning paradigm that contains a central server and distributed clients, offers a natural solution for aligning LLMs using decentralized and private human preference data held by different clients.

\begin{figure}[t]
    \centering
    \includegraphics[width=0.48\textwidth]{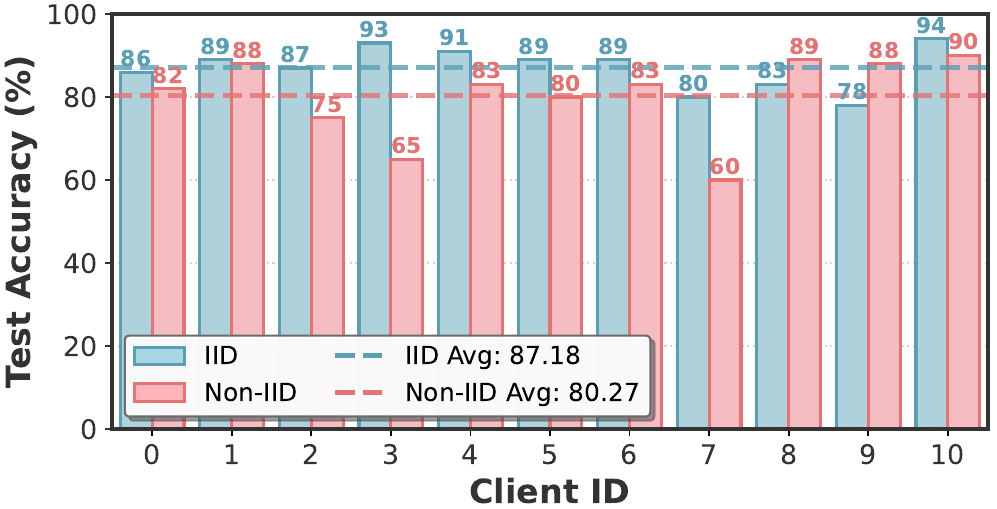}
       \vspace{-1.2em}
    \caption{Accuracy comparison of DPO under IID and non-IID settings on the code vulnerability dataset.
    \vspace{-1em}
    }
    \label{fig:iid_vs_non_iid}
\end{figure}

Direct Preference Optimization (DPO) \citep{rafailov2023direct} is a simple and effective alternative to reinforcement learning with human feedback (RLHF) \citep{ouyang2022training,bai2022constitutional,schulman2017proximal} for aligning LLMs with human preferences, as it removes the need for explicit reward models and avoids costly online rollouts while achieving strong empirical performance \citep{rafailov2023direct,tunstall2023zephyr}.
Therefore, DPO is a more suitable choice for aligning LLMs with human preferences in FL scenarios, as it does not require aggregating explicit reward models across clients and eliminates the need for online rollouts, thereby reducing both computational and communication overhead \citep{ouyang2022training}. 
Although a recent work, {\tt{FedDPO}} \citep{ye2024openfedllm}, has demonstrated the feasibility of incorporating DPO into {\tt{FedAvg}} (a canonical FL algorithm) for federated preference alignment, two significant challenges remain unaddressed:
\begin{itemize}
    \item \textbf{Non-IID data heterogeneity.} Figure~\ref{fig:iid_vs_non_iid} shows that, in our preliminary experiments, DPO suffers from significant performance degradation when trained on non–independent and identically distributed (non-IID) preference data compared to the IID setting. However, in practical FL scenarios, preference data across different clients are typically non-IID \citep{zhao2018federated}.
    
    \item \textbf{Limited generalization of implicit rewards.} DPO, which relies solely on implicit reward estimation, may exhibit poor generalization when faced with highly diverse preference distributions \citep{lin2024limited,li2023policy,yang2024regularizing, azar2024general}, further deteriorating alignment quality in FL with non-IID preference data.
\end{itemize}

To align large language models (LLMs) with human preferences in FL scenarios while tackling the aforementioned challenges, we propose {\tt{FedPDPO}} (\underline{Fed}erated \underline{P}ersonalized \underline{D}irect \underline{P}reference \underline{O}ptimization).
First, we construct a federated LLM fine-tuning architecture for aligning LLMs with human preferences.
Specifically, each client maintains a frozen pretrained LLM augmented with a parallel Low-Rank Adaptation (LoRA) adapter \citep{hu2022lora}. During local training, only the LoRA adapter is fine-tuned locally and subsequently aggregated across clients on the server, significantly reducing both communication and computational overhead by avoiding the transmission and training of full LLM parameters.
Building on this architecture, we further introduce three key innovations that explicitly address the two challenges discussed above.

(i) To mitigate the performance degradation induced by non-IID data heterogeneity, we decompose the LLM into two components: a frozen backbone and a trainable task-specific head. Each client shares the fine-tuned LoRA adapter to capture global knowledge, while maintaining a personalized local LLM head to adapt to its non-IID data distribution.
This design is inspired by personalized federated learning (PFL) approaches \citep{kulkarni2020survey}, which combine partial parameter sharing with partial parameter personalization.               
(ii) To alleviate the limited generalization of implicit rewards in DPO, we propose a new personalized DPO training strategy, termed \textbf{PDPO}. In practice, we introduce a client-specific explicit reward head parallel to the personalized LLM head, which complements the implicit rewards used in DPO and also further alleviate non-IID issues.
(iii) We observe that the features integrated from the LLM backbone and the globally shared LoRA adapter predominantly capture global feature distributions while inadequately representing local characteristics, which may disrupt the balance between global and local knowledge.
To address this issue, we introduce a bottleneck adapter that serves as a fused feature space, bridging the LLM backbone and the LoRA adapter with two personalized heads.

Our main contributions can be summarized as follows:
\begin{itemize}

\item We propose {\tt{FedPDPO}}, the first framework for aligning LLMs with human preferences across clients in FL while preserving preference privacy through parameter-efficient fine-tuning and aggregation of LoRA adapters on frozen client LLM backbones.

\item We share the fine-tuned LoRA adapter and personalize the client-specific LLM head to mitigate non-IID heterogeneity. We design a client-specific explicit reward head and devise a personalized DPO training strategy to enhance generalization and further alleviate non-IID issues. We propose a bottleneck adapter as a feature fusion space to balance the global and local features.

\item We provide theoretical analysis establishing the probabilistic foundation and soundness of {\tt{FedPDPO}}, and conduct 
extensive experiments on multiple preference datasets, demonstrating state-of-the-art performance with up to a $4.80\%$ average accuracy improvement under federated intra- and cross-domain settings.
\end{itemize}

\section{Related Work}

\subsection{Personalized Federated Learning (PFL)} 

Personalized Federated Learning (PFL) \citep{tan2022towards} has been proposed to address data heterogeneity (\emph{i.e.}, non-IID issue) across clients and enable client-specific model adaptation in FL \citep{kulkarni2020survey,tan2022towards,arivazhagan2019federated,fallah2020personalized,collins2021exploiting}. Unlike conventional FL, which aims to learn a single global model, PFL seeks to learn personalized models for each client by jointly leveraging shared global knowledge and local data characteristics.

Existing PFL methods can be categorized into four classes. 
(a) Meta-learning-based methods, such as \texttt{Per-FedAvg} \citep{fallah2020personalized} and \texttt{FedMeta} \citep{chen2018federated}, fine-tune a global meta model to adapt to heterogeneous client data.
(b) Regularization-based methods, including \texttt{FedProx} \citep{li2020federated}, \texttt{pFedMe} \citep{t2020personalized}, and \texttt{Ditto} \citep{li2021ditto}, introduce regularization terms during local optimization to balance global consistency and client-specific adaptation.
(c) Personalized aggregation-based methods, such as \texttt{FedFomo} \citep{zhang2020personalized}, \texttt{FedAMP} \citep{huang2021personalized}, and \texttt{FedALA} \citep{zhang2023fedala}, construct personalized models by selectively aggregating or weighting models from other clients.

Most relevant to our work are (d) personalized-head-based methods, including \texttt{FedPer} \citep{arivazhagan2019federated} and \texttt{FedRep} \citep{collins2021exploiting}, which decompose the model into a globally shared feature extractor and a client-specific head. By sharing the feature extractor across clients while training the heads locally, these approaches effectively capture global features while allowing personalization for heterogeneous data distributions.

Unlike prior PFL methods that target supervised learning tasks, our work studies personalized federated preference alignment of LLMs. 
In contrast to (d) personalized-head-based PFL methods, which share a trainable backbone across clients, we decompose the LLM into a backbone and a personalized head, freeze the backbone, and aggregate only its LoRA adapters. This design enables parameter-efficient global knowledge sharing while preserving client-level personalization.

\subsection{DPO for LLM Alignment in FL} 

Reinforcement learning with human feedback (RLHF) has been widely adopted to align large language models (LLMs) with human preferences. Among existing approaches, proximal policy optimization (PPO) \citep{schulman2017proximal, ouyang2022training, stiennon2020learning, wang2020truly} is a commonly used policy optimization algorithm within RLHF, while direct preference optimization (DPO) has recently emerged as an efficient alternative that directly optimizes the policy from preference data without explicit reward models or online rollouts \citep{rafailov2023direct}. Compared to PPO-based RLHF, DPO significantly improves training efficiency and stability in centralized settings.

Recent studies have provided theoretical analyses on the convergence and generalization properties of DPO \citep{rafailov2023direct, xu2024dpo}. However, DPO relies on implicit reward estimation, which has been shown to generalize poorly under diverse or out-of-distribution preference distributions \citep{lin2024limited, li2023policy, yang2024regularizing, jia2024generalizing}. This limitation becomes more pronounced when preference data are highly heterogeneous.

In FL scenarios, preference data collected by decentralized clients are typically non-IID, making the direct incorporation of DPO particularly challenging. A recent work, {\tt{FedDPO}} \citep{ye2024openfedllm}, applies DPO within the standard {\tt{FedAvg}} framework, but does not explicitly address non-IID data heterogeneity or the generalization limitations of implicit rewards, leading to suboptimal performance.

In contrast, our {\tt{FedPDPO}} introduces a personalized federated DPO training strategy by incorporating a client-specific explicit reward head to complement implicit reward signals, which explicitly addresses both non-IID preference distributions and the generalization limitations of implicit rewards. 
To the best of our knowledge, {\tt{FedPDPO}} is the first work to  combine DPO with personalized federated learning for LLM alignment under heterogeneous preference distributions.

\begin{figure*}[t]
    \vspace{0.1em}
    \centering
    \includegraphics[width=0.9\textwidth]{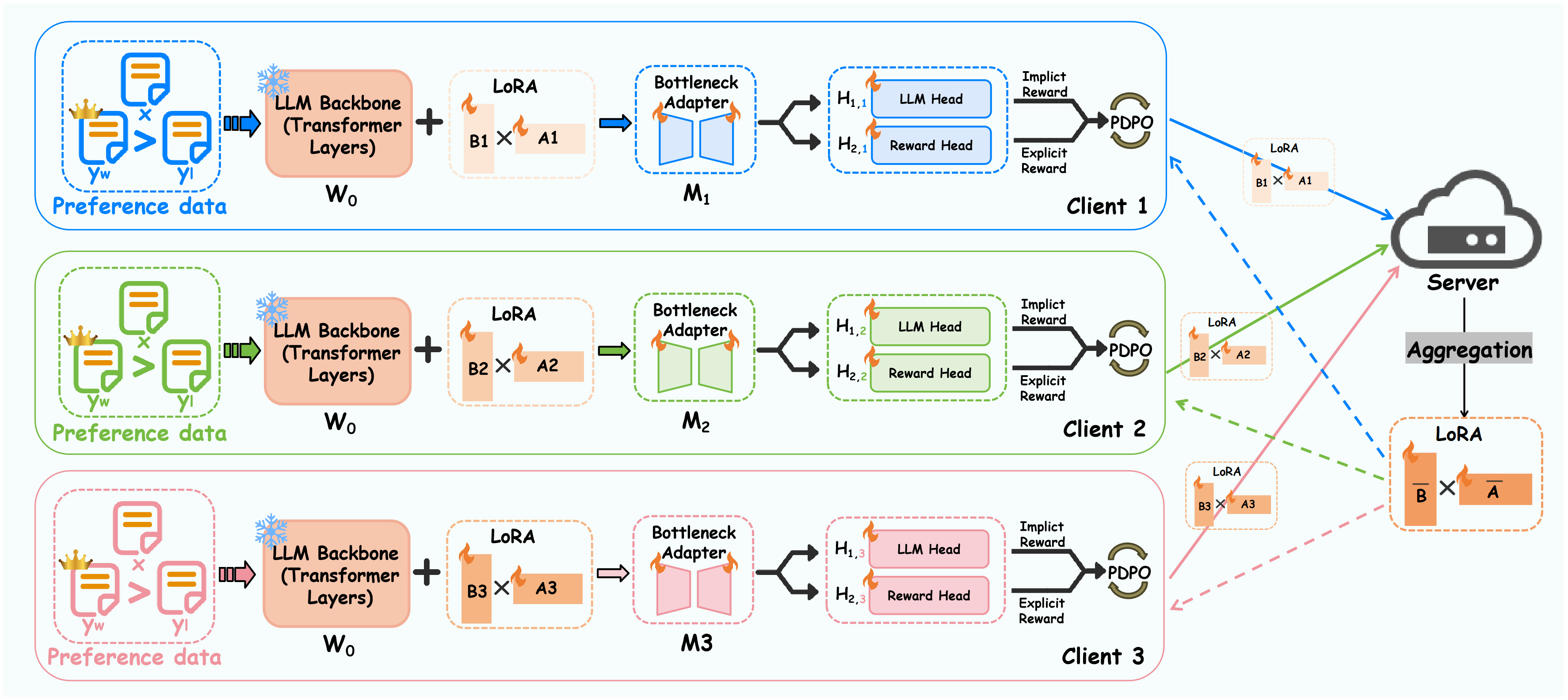}
    \vspace{0.7em}
    \caption{Overview of \texttt{FedPDPO}. 
    Each client fine-tunes and shares a LoRA adapter $(\boldsymbol{A},\boldsymbol{B})$ on a frozen backbone $\boldsymbol{W}_0$ while locally training personalized modules: a bottleneck adapter $\boldsymbol{M}$, local LLM and reward heads $(\boldsymbol{h}_1,\boldsymbol{h}_2)$ with an effective PDPO training strategy.}
    \label{fig:fedpdpo_framework}
    \vspace{-0.2em}
\end{figure*}

\section{Methodology}

\subsection{Problem Definition}

\textbf{Optimization Objective under Non-IID Preferences.}
We consider FL under data-heterogeneous settings.
Specifically, we assume there are $N$ clients, each holding a private local preference dataset $\{\mathcal{D}_1, \ldots, \mathcal{D}_N\}$ with distinct underlying data distributions and varying data amounts, \emph{i.e.},
non-independent and identically distributed (non-IID) and unbalanced.
Under the coordination of a central server, the objective of PFL is to collaboratively learn a collection of client-specific models $\{\boldsymbol{\Theta}_1^{*}, \ldots, \boldsymbol{\Theta}_N^{*}$\} by minimizing a global objective:
\begin{equation}
\{\boldsymbol{\Theta}_1^*,\ldots,\boldsymbol{\Theta}_N^*\}
\;=\;
\arg\min_{\boldsymbol{\Theta}_1,\ldots,\boldsymbol{\Theta}_N} \; 
\sum_{i=1}^{N} p_i \, \mathcal{L}_i(\boldsymbol{\Theta}_i,\mathcal{D}_i),
\label{eq:global_obj}
\end{equation}
where $\mathcal{L}_i(\boldsymbol{\Theta}_i, \mathcal{D}_i)$ denotes the loss of the client-specific model $\boldsymbol{\Theta}_i$ on the private preference data $\mathcal{D}_i$ of client $i$,
and $p_i = \tfrac{|\mathcal{D}_i|}{\sum_{j=1}^{N} |\mathcal{D}_j|}$ is the aggregation weight proportional to the size of the local dataset $|\mathcal{D}_i|$.

\textbf{Personalized Model Decomposition with LoRA Adapters.}
We decompose each client LLM $\boldsymbol{\Theta}_i$ into two components: a LLM backbone $\mathbf{W}_0$ and a client-specific LLM head $\mathbf{h}_{1,i}$. As the backbone $\mathbf{W}_0$ typically contains a large number of parameters, we adopt Low-Rank Adaptation (LoRA) as a parameter-efficient fine-tuning mechanism by injecting a LoRA adapter in parallel to the backbone. The LoRA adapter is parameterized by two low-rank matrices $(\mathbf{A}_i, \mathbf{B}_i)$. During local training, the backbone $\mathbf{W}_0$ is frozen, while the LoRA adapter is fine-tuned and shared across clients through aggregation to enable global knowledge exchange.

In parallel with the client-specific LLM head $\mathbf{h}_{1,i}$, we further introduce a personalized explicit reward head $\mathbf{h}_{2,i}$ to model client-specific preference signals. 
To balance global features learned by the shared LoRA adapter and the local features extracted by the frozen LLM backbone, we additionally design a bottleneck adapter $\mathbf{M}_i$ between the backbone (augmented with LoRA) and the two heads, which serves as a feature fusion space. Consequently, the complete client model $\boldsymbol{\Theta}_i$ can be formulated as:
\begin{equation}
\boldsymbol{\Theta}_i = \{(\mathbf{W}_0 \Vert\ (\mathbf{A}_i, \mathbf{B}_i)) \circ \mathbf{M}_{i} \circ (\mathbf{h}_{1,i} \Vert\ \mathbf{h}_{2,i})\}.
\label{eq:model_decomp}
\end{equation}

In each FL round, client $i$ freezes the LLM backbone $\mathbf{F}_i$ and updates the LoRA adapter $(\mathbf{A}_i,\mathbf{B}_i)$ together with the remaining personalized modules. The updated LoRA adapters are then aggregated across clients to facilitate global knowledge sharing, while the personalized modules remain local to adapt to client-specific preference distributions. This design achieves communication efficiency and effectively mitigates performance degradation under non-IID data.\footnote{\scriptsize Table.~\ref{tab:notation} (Appendix \ref{app:notation}) summarizes the key notations in {\tt{FedPDPO}}.}

\subsection{The FedPDPO Framework}
\label{sec:fedpdpo_framework}

Built upon the proposed federated parameter-efficient LLM architecture, {\tt FedPDPO} comprises three key methodological components that jointly tackle non-IID data heterogeneity and the limited generalization of implicit preference rewards:
(i) personalized federated parameter-efficient LLM fine-tuning, 
(ii) personalized DPO with client-specific reward modeling,
(iii) global-local feature fusion via {{bottleneck adapter}}.

\subsubsection{Personalized Federated Fine-Tuning}
We design a personalized federated LLM fine-tuning strategy that explicitly balances global knowledge sharing and client-specific adaptation under non-IID preference data. The core idea is to decouple the model into a shared, communication-efficient component and a personalized, locally adaptive component.

\textbf{Local Personalized Training.}
Concretely, each client LLM $\mathbf{W}_i$ is decomposed into a frozen LLM backbone $\mathbf{W}_0$ and a trainable client-specific LLM head $\mathbf{h}_{1,i}$. To enable efficient collaboration across clients without fine-tuning the full backbone, we adopt Low-Rank Adaptation (LoRA) \citep{hu2022lora} as a parameter-efficient fine-tuning mechanism. For a pretrained weight matrix $\mathbf{W}_0 \in \mathbb{R}^{m \times n}$ in the backbone, LoRA parameterizes the update as $\mathbf{W} = \mathbf{W}_0 + \mathbf{B}\mathbf{A}$, where $\mathbf{A} \in \mathbb{R}^{r \times n}$ and $\mathbf{B} \in \mathbb{R}^{m \times r}$ are trainable low-rank matrices with $r \ll \min(m,n)$. During local training, the backbone $\mathbf{W}_0$ remains frozen and only the LoRA adapters $(\mathbf{A}_i,\mathbf{B}_i)$ and the client-specific LLM head $\mathbf{h}_{1,i}$ are updated.

\textbf{Communication-efficient Aggregation.} 
The LoRA adapters act as the shared model components and are aggregated across clients to exchange global preference-related knowledge, while the client-specific LLM heads are trained locally and kept private, enabling each client to adapt the model outputs to its own non-IID preference distribution. 

By jointly fine-tuning the shared LoRA adapters and personalized LLM heads, \texttt{FedPDPO} achieves communication-efficient collaboration via low-rank parameter sharing while preserving client-level personalization, thereby effectively alleviating performance degradation caused by data heterogeneity in federated preference alignment.

\subsubsection{Personalized DPO Optimization}

To enhance the generalization of preference optimization under non-IID data, we design a personalized DPO training strategy that jointly leverages implicit preference signals from the language model head and explicit preference signals from a client-specific reward head.

\paragraph{Personalized Dual-Head Design.}
For each client $i$, we introduce a dual-head architecture. The LLM head $\mathbf{h}_{1,i}$ produces token-level logits, which define the conditional log-probability $\log \pi_\theta(y \mid x)$ used by standard DPO \citep{rafailov2023direct}. In parallel, we introduce a personalized explicit reward head $\mathbf{h}_{2,i}$ that directly estimates a scalar preference score for a given input--output pair $(x,y)$:
\begin{equation}
\mathbf{h}_{2,i}(x,y) = f_{\text{mlp}}(\textit{\text{Pool}}(\boldsymbol{z})),
\label{eq:reward_head}
\end{equation}
where $f_{\text{mlp}}(\cdot)$ is a two-layer MLP with \textit{Tanh} activation and \textit{dropout}, and $\textit{\text{Pool}}(\boldsymbol{z})$ denotes a sequence-level pooling operator on the feature $\boldsymbol{z}\in \mathbb{R}^d$ extracted by the piror layers. Both heads are trained locally and remain fully personalized, enabling adaptation to client-specific preference distributions and hence alleviating the non-IID challenge.

\paragraph{Implicit Preference Optimization via DPO.}
Given a preference tuple\footnote{\scriptsize We give an example of a preference tuple in {{Appendix \ref{app:data_format}}}.} $(x, y_w, y_l)$ consisting of an input prompt $x$, a preferred response $y_w$, and a less preferred response $y_l$, standard DPO defines an implicit reward margin:
\begin{equation}
\Delta_{\text{ir}}
=
\log \frac{\pi_\theta(y_w \mid x)}{\pi_{\text{ref}}(y_w \mid x)}
-
\log \frac{\pi_\theta(y_l \mid x)}{\pi_{\text{ref}}(y_l \mid x)},
\label{eq:dpo_margin_fraction}
\end{equation}
where $\pi_{\text{ref}}$ denotes a frozen reference policy. The corresponding objective minimizes a logistic loss over the implicit margin, encouraging the model to assign higher likelihood to preferred responses.

\paragraph{PDPO with Explicit Reward Augmentation.}
While effective, the implicit reward in DPO may generalize poorly under highly heterogeneous preference distributions. To address this limitation, we augment DPO with an explicit reward margin computed by the personalized reward head:
\begin{equation}
\Delta_{\text{er}}
=
\mathbf{h}_{2,i}(x,y_w)
-
\mathbf{h}_{2,i}(x,y_l).
\label{eq:reward_margin}
\end{equation}
The implicit and explicit margins are combined to form a unified preference signal:
\begin{equation}
\Delta
=
\Delta_{\text{ir}}
+
w_r \cdot s \cdot \Delta_{\text{er}},
\label{eq:combined_margin}
\end{equation}
where $w_r \ge 0$ controls the contribution of the explicit reward, and $s$ is an adaptive scaling factor that aligns the magnitudes of the two signals.

\paragraph{Adaptive Scaling and PDPO Objective.}
To stabilize training, the scaling factor is computed at the batch level as
\begin{equation}
s
=
\frac{\mathbb{E}[|\Delta_{\text{ir}}|]}{\mathbb{E}[|\Delta_{\text{er}}|] + \epsilon},
\label{eq:adaptive_scale}
\end{equation}
with exponential moving average smoothing and a small constant $\epsilon$ for numerical stability. The reward weight $w_r$ is linearly scheduled across communication rounds, allowing the explicit reward signal to gradually assume greater influence as training progresses.

The final PDPO objective retains the logistic form of DPO but optimizes the combined margin:
\begin{equation}
\mathcal{L}_{\text{PDPO}}
=
-\mathbb{E}_{(x,y_w,y_l)}\big[
\log \sigma\!\big(\beta \cdot \Delta\big)
\big],
\label{eq:pdpo_loss}
\end{equation}
where $\sigma$ is the sigmoid function and $\beta > 0$ is a temperature
that controls the deviation from the frozen reference policy.

By jointly optimizing implicit and explicit preference signals through personalized dual heads, PDPO improves robustness and generalization under non-IID preference distributions, providing a principled foundation for personalized federated preference alignment.

\begin{algorithm}[tb]
    \caption{\texttt{FedPDPO} Algorithm}
    \begin{algorithmic}[1]
        \STATE \textbf{Initialize:} 
        \STATE \quad {Server}: global LoRA adapter $(\mathbf{A}^0, \mathbf{B}^0)$;
        \STATE \quad {Each client $i$}: personalized modules $\mathbf{M}_i^0, \mathbf{h}_{1,i}^0, \mathbf{h}_{2,i}^0$;
        \STATE \quad Hyperparameters: $\eta_h, \eta_w$;
        \STATE \quad Maximum number of rounds: $T$.
        \FOR{each round $t = 1, \ldots, T$}
            \STATE \textbf{Server} broadcasts $(\mathbf{A}^{t-1}, \mathbf{B}^{t-1})$ to all clients.
            \FOR{each client $i \in \{1,\ldots,N\}$ \textbf{in parallel}}
                \STATE $(\mathbf{A}_i^{t-1}, \mathbf{B}_i^{t-1}) \leftarrow (\mathbf{A}^{t-1}, \mathbf{B}^{t-1})$;
                \STATE $(\mathbf{M}_i^{t}, \mathbf{h}_{1,i}^{t}, \mathbf{h}_{2,i}^{t}) \leftarrow (\mathbf{M}_i^{t-1}, \mathbf{h}_{1,i}^{t-1}, \mathbf{h}_{2,i}^{t-1})$.
                 \STATE /* Local alternative training: */
                \STATE  (1) Freeze LoRA, train personalized modules:
                \STATE \ $\mathbf{M}_i^{t} \leftarrow \mathbf{M}_i^{t-1} - \eta_h \nabla_{\mathbf{M}_i} \mathcal{L}_{\text{PDPO}}$;
                \STATE \ $\mathbf{h}_{1,i}^{t} \leftarrow \mathbf{h}_{1,i}^{t-1} - \eta_h \nabla_{\mathbf{h}_{1,i}} \mathcal{L}_{\text{PDPO}}$;
                \STATE  \ $\mathbf{h}_{2,i}^{t} \leftarrow \mathbf{h}_{2,i}^{t-1} - \eta_h \nabla_{\mathbf{h}_{2,i}} \mathcal{L}_{\text{PDPO}}$.
                \STATE (2) Freeze personalized modules, train LoRA:
                \STATE   $(\mathbf{A}_i^{t}, \mathbf{B}_i^{t}) \leftarrow (\mathbf{A}_i^{t-1}, \mathbf{B}_i^{t-1}) - \eta_w \nabla_{(\mathbf{A}_i,\mathbf{B}_i)} \mathcal{L}_{\text{PDPO}}$.
                \STATE Upload $(\mathbf{A}_i^{t}, \mathbf{B}_i^{t})$ to server.
            \ENDFOR
            \STATE \textbf{Server aggregation:}
            \STATE \quad $\mathbf{A}^{t} = \sum_{i=1}^N p_i \mathbf{A}_i^{t}$, $\mathbf{B}^{t} = \sum_{i=1}^N p_i \mathbf{B}_i^{t}$,
            \STATE \quad where $p_i = \frac{|\mathcal{D}_i|}{\sum_{j=1}^N |\mathcal{D}_j|}$, as defined in Eq.~\eqref{eq:global_obj}.
        \ENDFOR
    \end{algorithmic}
    \label{algorithm_fedpdpo}
\end{algorithm}

\paragraph{Alternating Optimization Strategy.}
During local training, we adopt an alternating optimization strategy to enable bidirectional knowledge transfer between global and personalized components, as shown in Alg. \ref{algorithm_fedpdpo}.
Specifically, we first freeze the shared LoRA adapters and update all personalized modules—including the bottleneck adapter and dual heads—to adapt to client-specific preference distributions. We then freeze the personalized modules and optimize the LoRA adapters, allowing locally learned preference signals to be distilled into the shared low-rank representations. This alternating scheme effectively balances global collaboration and local personalization under non-IID data.

\subsubsection{Global-Local Feature Fusion}
While the frozen LLM backbone preserves client-specific features from local data, the globally shared LoRA adapters capture transferable preference-related knowledge across clients. Directly feeding these two sources of features into task-specific heads, however, may lead to an imbalanced use of global and local signals under non-IID preference distributions. 

To address this issue, we introduce a lightweight bottleneck adapter $\mathbf{M}_i$ that explicitly fuses and balances local and global knowledge before task-specific prediction. Inserted between the backbone (augmented with LoRA adapters) and the personalized dual heads, the bottleneck adapter is jointly trained with both heads to adaptively reconcile frozen local backbone features with globally shared LoRA-induced features.

\paragraph{Bottleneck Adapter Design.}
Given the hidden state $\boldsymbol{z}^\prime \in \mathbb{R}^{d}$ produced by the LLM backbone with LoRA adaptation, the bottleneck adapter transforms the feature as
\begin{equation}
\boldsymbol{z} = \mathbf{M}_i(\boldsymbol{z}^\prime) =
\textit{LayerNorm}\big(
\boldsymbol{z}^\prime + f_{\cdot}(\boldsymbol{z}^\prime)
\big),
\label{eq:bottleneck_adapter}
\end{equation}
where
$
f_{\cdot}(\boldsymbol{z}^\prime)
=
\textit{Decode}\!\left(
\textit{Dropout}\!\left(
\textit{GELU}\!\left(
\textit{Encode}(\boldsymbol{z}^\prime)
\right)
\right)
\right)
$
implements a bottleneck transformation with a low-dimensional intermediate space. This residual bottleneck structure allows the adapter to selectively modulate and reweight feature dimensions, rather than overwriting the original features.

\paragraph{Balancing Global and Local Knowledge.}
The bottleneck adapter integrates global preference information encoded by the shared LoRA adapters with local characteristics preserved in the frozen backbone. By operating in a low-dimensional feature space, the adapter adaptively controls the influence of global and local signals, preventing domination by globally shared features and improving robustness to data heterogeneity.

As a result, the personalized LLM head and explicit reward head receive balanced and client-adaptive features, which further enhances personalization and stability in federated preference optimization.

\section{Theoretical Analysis of PDPO}
\label{sec:theory}

We provide a concise theoretical justification for PDPO by interpreting the explicit reward head as an additive correction under a random-utility preference model.

\paragraph{Theorem 1 (Gumbel--Bradley--Terry equivalence).}
Let $y_w$ and $y_l$ be two responses to a prompt $x$, and let $\mathbf{h}_{2,i}(x,y_w)$ and $\mathbf{h}_{2,i}(x,y_l)$ denote their scores produced by the reward head. Define
\[
\begin{aligned}
R_w &\sim \operatorname{Gumbel}\!\left(r_\theta(x,y_w),\,1\right), \\
R_l &\sim \operatorname{Gumbel}\!\left(r_\theta(x,y_l),\,1\right),
\end{aligned}
\]
where $R_w$ and $R_l$ are independent. Then,
\[
\Pr(R_w - R_l > 0)
= p_{BT}(y_w \succ y_l \mid x)
= \sigma(\Delta_{\text{er}}),
\]
with $\Delta_{\text{er}} = \mathbf{h}_{2,i}(x,y_w) - \mathbf{h}_{2,i}(x,y_l)$.

\noindent\textbf{Proof.}
By the Gumbel--max trick~\citep{maddison2016concrete}. See Appendix~\ref{app:theorem1_proof}.

\paragraph{Theorem 2 (Reward-corrected comparison).}
For any scalar $c \in \mathbb{R}$,
\[
\Pr(R_w - R_l + c > 0)
= \sigma(\Delta_{\text{er}} + c).
\]

\noindent\textbf{Proof.}
The detailed proof is provided in Appendix~\ref{app:theorem2_proof}.

\paragraph{Implication for PDPO.}
Theorem~2 shows that adding a correction term $c$ shifts the Bradley--Terry decision boundary without changing its functional form. Consequently, PDPO corresponds to maximum-likelihood estimation under a Bradley--Terry model with a corrected margin. When $c = w_r \cdot s \cdot \Delta_{\text{er}}$, the explicit reward head provides an adaptive additive correction to the implicit reward, yielding the PDPO objective.

\section{Experiments}
We conduct experiments on an NVIDIA GeForce RTX 3090 GPU with 24\,GB memory and simulate the federated learning environment by sequentially executing multiple clients.


\textbf{Datasets.}
We evaluate \texttt{FedPDPO} under two federated settings: \emph{intra-domain} and \emph{cross-domain}. 
In the intra-domain setting, all clients are trained on the same preference dataset, including \textbf{IMDB} \citep{maas2011learning}\footnote{\tiny\url{https://huggingface.co/datasets/yuasosnin/imdb-dpo}} and \textbf{Code-Vulnerability-Security}\footnote{\tiny\url{https://huggingface.co/datasets/CyberNative/Code_Vulnerability_Security_DPO}}. To simulate data heterogeneity within a shared domain, we further construct non-IID client partitions using reward-margin–based and language-based splitting strategies. 
In the cross-domain setting, each client is assigned a distinct preference dataset—\textbf{WebGPT} \citep{nakano2021webgpt}\footnote{\tiny\url{https://huggingface.co/datasets/openai/webgpt_comparisons}}, \textbf{PyDPO}\footnote{\tiny\url{https://huggingface.co/datasets/jondurbin/py-dpo-v0.1}}, or \textbf{UltraFeedback} \citep{cui2023ultrafeedback}\footnote{\tiny\url{https://huggingface.co/datasets/kaitchup/UltraFeedback-prompt-chosen-rejected}}—to model highly diverse human feedback sources across domains (open-ended question answering, code generation, and instruction following).
The more details are provided in Appendix~\ref{app:datasets}.

\textbf{Models.}
Unless otherwise specified, we use \textbf{DistilGPT2} \citep{sanh2019distilbert}\footnote{\tiny\url{https://huggingface.co/distilbert/distilgpt2}} as the backbone LLM ($d=768$). Clients share LoRA-adapted backbone parameters with rank $r=8$, while the backbone remains frozen. 
The bottleneck adapter projects features to a 256-dimensional latent space and back to $d=768$ with a residual connection and layer normalization. 
Each client maintains two personalized heads: an LLM head for computing token-level log-probabilities and a reward head implemented as a two-layer MLP (hidden size 256) that outputs scalar preference scores via mean pooling.
To assess generalization across model architectures, we additionally evaluate \texttt{FedPDPO} with \textbf{TinyLlama-1.1B }\citep{zhang2024tinyllama}\footnote{\tiny\url{https://github.com/jzhang38/TinyLlama}}.
More details are provided in Appendix~\ref{app:models}.

\textbf{Baselines.}
We compare \texttt{FedPDPO} with five representative federated learning methods: 
\texttt{FedAvg}~\citep{mcmahan2017communication} (which reduces to \texttt{FedDPO}~\citep{ye2024openfedllm} when combined with DPO), 
\texttt{Per-FedAvg}~\citep{fallah2020personalized}, 
\texttt{FedAMP}~\citep{huang2021personalized}, 
\texttt{FedPer}~\citep{arivazhagan2019federated}, and 
\texttt{FedRep}~\citep{collins2021exploiting}. 
These baselines cover standard FL, meta-learning–based personalization, attentive aggregation, and personalized-head approaches.

For baselines, we evaluate two preference optimization objectives: (\textit{i})  \textbf{PPO}, which relies on a frozen pre-trained reward model (deberta-v3-large-v2 \citep{kopf2023openassistant}\footnote{\tiny\url{https://huggingface.co/OpenAssistant/reward-model-deberta-v3-large-v2}} ), and (\textit{ii}) \textbf{DPO}, which directly optimizes policies from preference pairs without an explicit reward model.


\textbf{Evaluation Metrics.}
We use \emph{preference accuracy} as the primary evaluation metric, defined as the fraction of test samples for which the trained policy correctly ranks the preferred response over the rejected one. For each test triple $(x, y_w, y_l)$, we compute the log-likelihoods of both responses and count the prediction as correct if $\log p_\theta(y_w \mid x) > \log p_\theta(y_l \mid x)$. All results are averaged over five independent runs with different seeds.

\textbf{Hyperparameter Settings.}
We use \textit{AdamW} \citep{loshchilov2017decoupled} with cosine learning rate decay for all methods, selecting the learning rate from $\{1\times10^{-4},\,8\times10^{-5},\,5\times10^{-5},\,3\times10^{-5},\,1\times10^{-3}\}$. 
For PPO-based methods, the KL coefficient is set to $0.1$ with clipping parameter $0.2$.
For DPO-based methods, $\beta \in \{0.5,\ 0.1,\,0.05,\,0.01\}$. 
For \texttt{FedPDPO}, $w_r$ is linearly increased from $0.5$ to $1.5$ during training. 
Additional details and hyperparameter analysis are provided in Appendix~\ref{app:hyperparameters} and Appendix~\ref{app:extended_results}.

\subsection{Results on Intra-Domain Setting}

\label{sec:exp_homo}
Table~\ref{tab:compare-results} reports the intra-domain results under two non-IID client partitions with varying numbers of clients. 
On the IMDB dataset with reward-margin–based splits, \texttt{FedPDPO} achieves 87.41\% (3 clients) and 85.23\% (10 clients), outperforming the strongest baseline by +4.01\% and +4.80\%, respectively. 
On the Code-Vulnerability-Security dataset with language-based splits, \texttt{FedPDPO} further attains 96.92\% (3 clients) and 92.80\% (11 clients), yielding gains of +1.71\% and +4.55\% over the strongest baselines\footnote{\scriptsize The single accuracy of each client is provided in Table.~\ref{tab:homogeneous_results} (Appendix~\ref{app:extended_results}).}. 
These consistent improvements across both reward-margin and domain-specific heterogeneity demonstrate that \texttt{FedPDPO} effectively mitigates non-IID effects while scaling robustly with the number of clients. 

\begin{figure}[t]
  \centering
  \includegraphics[width=.8\columnwidth]{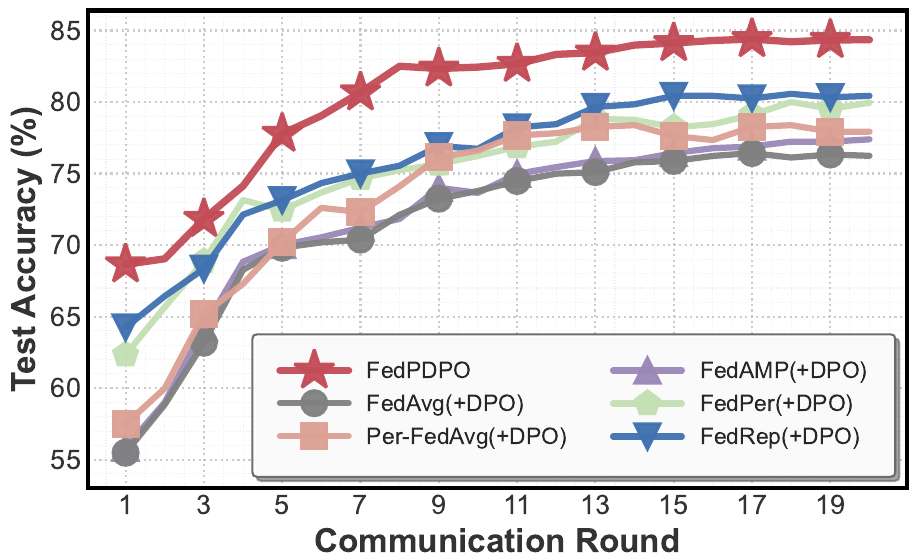}
  \caption{{{Test accuracy varies as communication rounds in \textit{intra-domain} FL settings with the IMDB dataset and 10 clients.}}}
  \label{fig:accuracy_curves_intra}
   \vspace{-1em}
\end{figure}

Across all settings, DPO-based methods consistently outperform their PPO counterparts, highlighting the superior efficiency and stability of DPO for federated preference alignment.
{{Figure~\ref{fig:accuracy_curves_intra}}} demonstrates that \texttt{FedPDPO} achieves faster convergence than baselines in intra-domains.

\begin{figure}[t]
  \centering
  \includegraphics[width=.8\columnwidth]{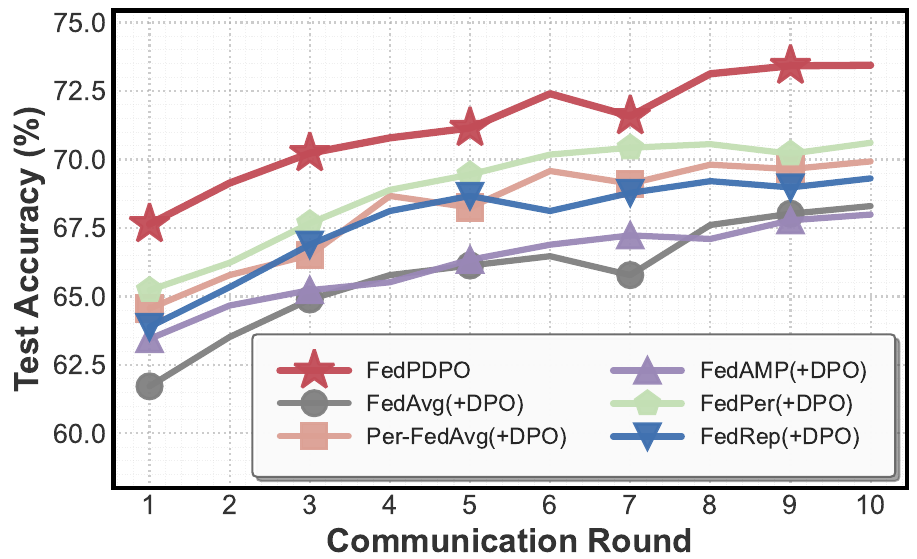}
  \caption{{{Test accuracy varies as communication rounds in \textit{cross-domain} FL settings with 3 domains of datasets assigned to 3 clients.}}}
  \label{fig:accuracy_curves_cross}
   \vspace{-1em}
\end{figure}

\begin{table*}[t]
\centering
\small
\caption{Test accuracy (\%) under intra-domain settings with 3, 10, or 11 clients and cross-domain settings with distinct datasets per client (3 clients). 
Code-V.: Code-Vulnerability-Security, UltraF.: UltraFeedback. 
\vspace{-0.7em}
\textbf{Bold} and \underline{underline} denote the best and second-best results.}
  \label{tab:compare-results}
\setlength{\tabcolsep}{5.8pt}
\renewcommand{\arraystretch}{1.08}
\begin{tabular}{lcccccccc}
\toprule
\multirow{2}{*}{\textbf{Method}} 
& \multicolumn{4}{c}{\textbf{Intra-Domain}} 
& \multicolumn{4}{c}{\textbf{Cross-Domain}} \\
\cmidrule(lr){2-5} \cmidrule(lr){6-9}
& \textbf{IMDB (3)} & \textbf{IMDB (10)} & \textbf{Code-V. (3)} & \textbf{Code-V. (11)} 
& \textbf{WebGPT} & \textbf{PyDPO} & \textbf{UltraF.} & \textbf{Avg.} \\
\midrule
FedAvg$_{\text{(+PPO)}}$ 
& 80.70$_{\pm0.24}$ & 70.89$_{\pm0.28}$ & 91.40$_{\pm0.20}$ & 78.32$_{\pm0.24}$ 
& 47.93$_{\pm0.22}$ & 80.12$_{\pm0.15}$ & 63.89$_{\pm0.29}$ & 63.98 \\
FedAvg$_{\text{(+DPO)}}$ 
& 82.73$_{\pm0.22}$ & 76.45$_{\pm0.23}$ & 93.49$_{\pm0.17}$ & 84.09$_{\pm0.19}$ 
& 52.92$_{\pm0.16}$ & 85.53$_{\pm0.09}$ & 66.45$_{\pm0.24}$ & 68.30 \\
\midrule
Per-FedAvg$_{\text{(+PPO)}}$ 
& 81.36$_{\pm0.28}$ & 73.12$_{\pm0.27}$ & 92.11$_{\pm0.24}$ & 79.68$_{\pm0.25}$ 
& 49.84$_{\pm0.28}$ & 81.76$_{\pm0.24}$ & 64.67$_{\pm0.32}$ & 65.42 \\
Per-FedAvg$_{\text{(+DPO)}}$ 
& \underline{83.40}$_{\pm0.25}$ & 78.83$_{\pm0.22}$ & 94.21$_{\pm0.21}$ & 85.44$_{\pm0.19}$ 
& 54.96$_{\pm0.23}$ & 87.36$_{\pm0.18}$ & 67.48$_{\pm0.27}$ & 69.93 \\
\midrule
FedAMP$_{\text{(+PPO)}}$ 
& 80.41$_{\pm0.30}$ & 71.45$_{\pm0.29}$ & 91.20$_{\pm0.26}$ & 81.23$_{\pm0.23}$ 
& 49.01$_{\pm0.26}$ & 78.43$_{\pm0.27}$ & 63.98$_{\pm0.35}$ & 63.81 \\
FedAMP$_{\text{(+DPO)}}$ 
& 82.75$_{\pm0.27}$ & 77.39$_{\pm0.24}$ & 93.38$_{\pm0.23}$ & 87.12$_{\pm0.17}$ 
& 54.58$_{\pm0.21}$ & 83.93$_{\pm0.22}$ & 65.47$_{\pm0.30}$ & 67.99 \\
\midrule
FedPer$_{\text{(+PPO)}}$ 
& 80.68$_{\pm0.32}$ & 75.01$_{\pm0.24}$ & 93.52$_{\pm0.23}$ & 80.54$_{\pm0.21}$ 
& 50.12$_{\pm0.25}$ & 83.21$_{\pm0.25}$ & 65.45$_{\pm0.33}$ & 66.26 \\
FedPer$_{\text{(+DPO)}}$ 
& 81.66$_{\pm0.29}$ & 79.95$_{\pm0.19}$ & \underline{95.21}$_{\pm0.20}$ & 86.37$_{\pm0.15}$ 
& \underline{55.38}$_{\pm0.20}$ & \underline{88.42}$_{\pm0.19}$ & \underline{68.02}$_{\pm0.28}$ & \underline{70.61} \\
\midrule
FedRep$_{\text{(+PPO)}}$ 
& 80.09$_{\pm0.31}$ & 74.58$_{\pm0.31}$ & 92.86$_{\pm0.25}$ & 82.15$_{\pm0.19}$ 
& 48.76$_{\pm0.27}$ & 81.09$_{\pm0.26}$ & 65.54$_{\pm0.34}$ & 65.13 \\
FedRep$_{\text{(+DPO)}}$ 
& 82.18$_{\pm0.28}$ & \underline{80.43}$_{\pm0.26}$ & 95.00$_{\pm0.22}$ & \underline{88.25}$_{\pm0.13}$ 
& 53.96$_{\pm0.22}$ & 86.71$_{\pm0.21}$ & 67.25$_{\pm0.29}$ & 69.31 \\
\midrule
\rowcolor{gray!10}
\textbf{\texttt{FedPDPO}}
& \textbf{87.41}$_{\pm0.20}$ & \textbf{85.23}$_{\pm0.14}$ & \textbf{96.92}$_{\pm0.15}$ & \textbf{92.80}$_{\pm0.13}$ 
& \textbf{58.92}$_{\pm0.14}$ & \textbf{91.47}$_{\pm0.12}$ & \textbf{69.92}$_{\pm0.19}$ & \textbf{73.44} \\
\bottomrule
\end{tabular}
\vspace{-1em}
\end{table*}

\subsection{Results on Cross-Domain Setting}

Table~\ref{tab:compare-results} reports the results in the cross-domain setting, where each client is trained on a preference dataset from a distinct domain. 
Despite this extreme non-IID scenario,
\texttt{FedPDPO} achieves the highest average accuracy of 73.44\%, beating the best baseline by +2.83\%. 
Figure~\ref{fig:accuracy_curves_cross} shows that \texttt{FedPDPO} also achieves faster convergence in cross-domains.


Notably, the improvements are consistent across all three domains, with gains of +3.54\% on WebGPT, +3.05\% on PyDPO, and +1.90\% on UltraFeedback compared to the strongest baseline \texttt{FedPer+DPO}. 
These results indicate that the reward-corrected dual-head design effectively captures both shared and domain-specific preference signals, mitigating the limited generalization of implicit rewards under severe distribution shifts. 

Consistent with the intra-domain results, DPO-based methods generally outperform PPO-based counterparts, highlighting DPO's robustness for cross-domain federated preference alignment.
Table \ref{tab:heterogeneous_tinyllama} shows that \texttt{FedPDPO} also consistently achieves the best test accuracy under cross-domain settings with the TinyLlama-1.1B model. 

These cross-domain gains are particularly significant given the heterogeneity across WebGPT (factual grounding), PyDPO (code quality), and UltraFeedback (instruction following). The improvements demonstrate that \texttt{FedPDPO} captures domain-invariant preference principles while enabling cross-domain complementarity, showcasing federated preference alignment's advantage over isolated training.

\begin{table}[t]
\centering
\small
\caption{Test accuracy (\%) under cross-domain settings with TinyLlama-1.1B on distinct datasets per client (3 clients).}
\vspace{-0.75em}
\label{tab:heterogeneous_tinyllama}
\setlength{\tabcolsep}{2.5pt}
\renewcommand{\arraystretch}{1.08}
\begin{tabular}{lcccc}
\toprule
\textbf{Method} & \textbf{WebGPT} & \textbf{PyDPO} & \textbf{UltraF.} & \textbf{Avg.} \\
\midrule
FedAvg$_{\text{(+PPO)}}$ & 50.85$_{\pm0.24}$ & 89.12$_{\pm0.19}$ & 69.54$_{\pm0.21}$ & 69.84 \\
FedAvg$_{\text{(+DPO)}}$ & 54.12$_{\pm0.18}$ & 91.80$_{\pm0.12}$ & 71.28$_{\pm0.22}$ & 72.40 \\
\midrule
Per-FedAvg$_{\text{(+PPO)}}$ & 52.34$_{\pm0.27}$ & 89.86$_{\pm0.22}$ & 70.15$_{\pm0.25}$ & 70.78 \\
Per-FedAvg$_{\text{(+DPO)}}$ & 55.67$_{\pm0.21}$ & 92.38$_{\pm0.17}$ & 72.41$_{\pm0.23}$ & 73.49 \\
\midrule
FedAMP$_{\text{(+PPO)}}$ & 51.58$_{\pm0.25}$ & 88.45$_{\pm0.24}$ & 69.12$_{\pm0.28}$ & 69.72 \\
FedAMP$_{\text{(+DPO)}}$ & 54.92$_{\pm0.20}$ & 91.23$_{\pm0.19}$ & 71.56$_{\pm0.24}$ & 72.57 \\
\midrule
FedPer$_{\text{(+PPO)}}$ & 53.18$_{\pm0.23}$ & 90.54$_{\pm0.21}$ & 71.38$_{\pm0.26}$ & 71.70 \\
FedPer$_{\text{(+DPO)}}$ & \underline{56.43}$_{\pm0.19}$ & \underline{93.05}$_{\pm0.16}$ & \underline{73.12}$_{\pm0.22}$ & \underline{74.20} \\
\midrule
FedRep$_{\text{(+PPO)}}$ & 52.47$_{\pm0.26}$ & 89.71$_{\pm0.23}$ & 70.29$_{\pm0.27}$ & 70.82 \\
FedRep$_{\text{(+DPO)}}$ & 55.81$_{\pm0.22}$ & 92.54$_{\pm0.18}$ & 72.67$_{\pm0.24}$ & 73.67 \\
\midrule
\rowcolor{gray!10}
\textbf{\texttt{FedPDPO}}
& \textbf{61.24$_{\pm0.16}$} & \textbf{94.32$_{\pm0.11}$} & \textbf{74.18$_{\pm0.17}$} & \textbf{76.58} \\
\bottomrule
\end{tabular}
 \vspace{-1.0em}
\end{table}


\subsection{Ablation Study}
We conduct ablation studies on the {IMDB} dataset with {10 clients} under reward-margin–based non-IID partitioning to assess the effect of each component in \texttt{FedPDPO}. Results are reported as the average test accuracy across clients.

\begin{table}[t]
  \centering
  \small
  \caption{Ablation study of \texttt{FedPDPO} on the IMDB dataset with 10 clients under reward-margin–based non-IID partitioning.}
  \label{tab:ablation}
  \setlength{\tabcolsep}{4pt}
  \renewcommand{\arraystretch}{1.1}
  \resizebox{0.4\textwidth}{!}{%
  \begin{tabular}{lc}
    \toprule
    \textbf{Case} & \textbf{Accuracy (\%)} \\
    \midrule
    (\textbf{A1}): w/o Bottleneck Adapter & 82.44$_{\pm0.18}$ \\
    (\textbf{A2}): w/o Reward Head & 81.94$_{\pm0.21}$ \\
    \rowcolor{gray!10}
    (\textbf{A3}): Full \texttt{FedPDPO} & \textbf{85.23$_{\pm0.14}$} \\
    \bottomrule
  \end{tabular}}
  \vspace{-1em}
\end{table}

\textbf{Ablation Settings.}
We evaluate the following variants:  
(\textbf{A1}) \emph{w/o Bottleneck Adapter}: removing the bottleneck adapter and directly feeding backbone features to both heads;  
(\textbf{A2}) \emph{w/o Reward Head}: removing the explicit reward head, reducing the method to standard DPO;  
(\textbf{A3}) \emph{Full \texttt{FedPDPO}}: the complete model with all components enabled.

\textbf{Results and Analysis.}
Removing the reward head (\textbf{A2}) leads to the largest performance drop, indicating that explicit reward signals are crucial for improving generalization under non-IID federated settings. 
Eliminating the bottleneck adapter (\textbf{A1}) yields moderate improvements over (\textbf{A2}) but still underperforms the full model, suggesting that feature fusion and balancing play an important role. 
The full \texttt{FedPDPO} (\textbf{A3}) consistently achieves the best accuracy, confirming the complementary benefits of the bottleneck adapter and the explicit reward correction.

\section{Conclusion}
We proposed \texttt{FedPDPO}, a personalized federated framework built on communication-efficient LoRA-based fine-tuning. By jointly personalizing local heads to address non-IID data, introducing an explicit reward head to design a new DPO training objective, and employing a bottleneck adapter for global--local feature fusion, achieving significant improvements in both intra- and cross-domain settings.

\newpage


\bibliographystyle{icml2026}
\bibliography{refs}           

@article{rafailov2023direct,
  title={Direct preference optimization: Your language model is secretly a reward model},
  author={Rafailov, Rafael and Sharma, Archit and Mitchell, Eric and Manning, Christopher D and Ermon, Stefano and Finn, Chelsea},
  journal={Advances in neural information processing systems},
  volume={36},
  pages={53728--53741},
  year={2023}
}

@article{ouyang2022training,
  title={Training language models to follow instructions with human feedback},
  author={Ouyang, Long and Wu, Jeffrey and Jiang, Xu and Almeida, Diogo and Wainwright, Carroll and Mishkin, Pamela and Zhang, Chong and Agarwal, Sandhini and Slama, Katarina and Ray, Alex and others},
  journal={Advances in neural information processing systems},
  volume={35},
  pages={27730--27744},
  year={2022}
}

@article{bai2022constitutional,
  title={Constitutional ai: Harmlessness from ai feedback},
  author={Bai, Yuntao and Kadavath, Saurav and Kundu, Sandipan and Askell, Amanda and Kernion, Jackson and Jones, Andy and Chen, Anna and Goldie, Anna and Mirhoseini, Azalia and McKinnon, Cameron and others},
  journal={arXiv preprint arXiv:2212.08073},
  year={2022}
}

@article{hu2022lora,
  title={Lora: Low-rank adaptation of large language models.},
  author={Hu, Edward J and Shen, Yelong and Wallis, Phillip and Allen-Zhu, Zeyuan and Li, Yuanzhi and Wang, Shean and Wang, Lu and Chen, Weizhu and others},
  journal={ICLR},
  volume={1},
  number={2},
  pages={3},
  year={2022}
}

@article{tan2022towards,
  title={Towards personalized federated learning},
  author={Tan, Alysa Ziying and Yu, Han and Cui, Lizhen and Yang, Qiang},
  journal={IEEE transactions on neural networks and learning systems},
  volume={34},
  number={12},
  pages={9587--9603},
  year={2022},
  publisher={IEEE}
}

@article{chen2018federated,
  title={Federated meta-learning with fast convergence and efficient communication},
  author={Chen, Fei and Luo, Mi and Dong, Zhenhua and Li, Zhenguo and He, Xiuqiang},
  journal={arXiv preprint arXiv:1802.07876},
  year={2018}
}

@inproceedings{kulkarni2020survey,
  title={Survey of personalization techniques for federated learning},
  author={Kulkarni, Viraj and Kulkarni, Milind and Pant, Aniruddha},
  booktitle={2020 fourth world conference on smart trends in systems, security and sustainability (WorldS4)},
  pages={794--797},
  year={2020},
  organization={IEEE}
}

@article{t2020personalized,
  title={Personalized federated learning with moreau envelopes},
  author={T Dinh, Canh and Tran, Nguyen and Nguyen, Josh},
  journal={Advances in neural information processing systems},
  volume={33},
  pages={21394--21405},
  year={2020}
}

@inproceedings{li2021ditto,
  title={Ditto: Fair and robust federated learning through personalization},
  author={Li, Tian and Hu, Shengyuan and Beirami, Ahmad and Smith, Virginia},
  booktitle={International conference on machine learning},
  pages={6357--6368},
  year={2021},
  organization={PMLR}
}

@article{li2020federated,
  title={Federated optimization in heterogeneous networks},
  author={Li, Tian and Sahu, Anit Kumar and Zaheer, Manzil and Sanjabi, Maziar and Talwalkar, Ameet and Smith, Virginia},
  journal={Proceedings of Machine learning and systems},
  volume={2},
  pages={429--450},
  year={2020}
}

@article{fallah2020personalized,
  title={Personalized federated learning with theoretical guarantees: A model-agnostic meta-learning approach},
  author={Fallah, Alireza and Mokhtari, Aryan and Ozdaglar, Asuman},
  journal={Advances in neural information processing systems},
  volume={33},
  pages={3557--3568},
  year={2020}
}

@article{maddison2016concrete,
  title={The concrete distribution: A continuous relaxation of discrete random variables},
  author={Maddison, Chris J and Mnih, Andriy and Teh, Yee Whye},
  journal={arXiv preprint arXiv:1611.00712},
  year={2016}
}

@inproceedings{maas2011learning,
  title={Learning word vectors for sentiment analysis},
  author={Maas, Andrew and Daly, Raymond E and Pham, Peter T and Huang, Dan and Ng, Andrew Y and Potts, Christopher},
  booktitle={Proceedings of the 49th annual meeting of the association for computational linguistics: Human language technologies},
  pages={142--150},
  year={2011}
}

@article{loshchilov2017decoupled,
  title={Decoupled weight decay regularization},
  author={Loshchilov, Ilya and Hutter, Frank},
  journal={arXiv preprint arXiv:1711.05101},
  year={2017}
}

@inproceedings{wang2020truly,
  title={Truly proximal policy optimization},
  author={Wang, Yuhui and He, Hao and Tan, Xiaoyang},
  booktitle={Uncertainty in artificial intelligence},
  pages={113--122},
  year={2020},
  organization={PMLR}
}

@article{nakano2021webgpt,
  title={Webgpt: Browser-assisted question-answering with human feedback},
  author={Nakano, Reiichiro and Hilton, Jacob and Balaji, Suchir and Wu, Jeff and Ouyang, Long and Kim, Christina and Hesse, Christopher and Jain, Shantanu and Kosaraju, Vineet and Saunders, William and others},
  journal={arXiv preprint arXiv:2112.09332},
  year={2021}
}

@article{cui2023ultrafeedback,
  title={Ultrafeedback: Boosting language models with scaled ai feedback},
  author={Cui, Ganqu and Yuan, Lifan and Ding, Ning and Yao, Guanming and He, Bingxiang and Zhu, Wei and Ni, Yuan and Xie, Guotong and Xie, Ruobing and Lin, Yankai and others},
  journal={arXiv preprint arXiv:2310.01377},
  year={2023}
}

@article{stiennon2020learning,
  title={Learning to summarize with human feedback},
  author={Stiennon, Nisan and Ouyang, Long and Wu, Jeffrey and Ziegler, Daniel and Lowe, Ryan and Voss, Chelsea and Radford, Alec and Amodei, Dario and Christiano, Paul F},
  journal={Advances in neural information processing systems},
  volume={33},
  pages={3008--3021},
  year={2020}
}

@article{achiam2023gpt,
  title={Gpt-4 technical report},
  author={Achiam, Josh and Adler, Steven and Agarwal, Sandhini and Ahmad, Lama and Akkaya, Ilge and Aleman, Florencia Leoni and Almeida, Diogo and Altenschmidt, Janko and Altman, Sam and Anadkat, Shyamal and others},
  journal={arXiv preprint arXiv:2303.08774},
  year={2023}
}

@article{singhal2023large,
  title={Large language models encode clinical knowledge},
  author={Singhal, Karan and Azizi, Shekoofeh and Tu, Tao and Mahdavi, S Sara and Wei, Jason and Chung, Hyung Won and Scales, Nathan and Tanwani, Ajay and Cole-Lewis, Heather and Pfohl, Stephen and others},
  journal={Nature},
  volume={620},
  number={7972},
  pages={172--180},
  year={2023},
  publisher={Nature Publishing Group}
}

@article{kasneci2023chatgpt,
  title={ChatGPT for good? On opportunities and challenges of large language models for education},
  author={Kasneci, Enkelejda and Se{\ss}ler, Kathrin and K{\"u}chemann, Stefan and Bannert, Maria and Dementieva, Daryna and Fischer, Frank and Gasser, Urs and Groh, Georg and G{\"u}nnemann, Stephan and H{\"u}llermeier, Eyke and others},
  journal={Learning and individual differences},
  volume={103},
  pages={102274},
  year={2023},
  publisher={Elsevier}
}

@article{schulman2017proximal,
  title={Proximal policy optimization algorithms},
  author={Schulman, John and Wolski, Filip and Dhariwal, Prafulla and Radford, Alec and Klimov, Oleg},
  journal={arXiv preprint arXiv:1707.06347},
  year={2017}
}

@article{li2020federated2,
  title={Federated learning: Challenges, methods, and future directions},
  author={Li, Tian and Sahu, Anit Kumar and Talwalkar, Ameet and Smith, Virginia},
  journal={IEEE signal processing magazine},
  volume={37},
  number={3},
  pages={50--60},
  year={2020},
  publisher={IEEE}
}

@article{kairouz2021advances,
  title={Advances and open problems in federated learning},
  author={Kairouz, Peter and McMahan, H Brendan},
  journal={Foundations and trends in machine learning},
  volume={14},
  number={1-2},
  pages={1--210},
  year={2021},
  publisher={Emerald Publishing Limited}
}

@article{zhao2018federated,
  title={Federated learning with non-iid data},
  author={Zhao, Yue and Li, Meng and Lai, Liangzhen and Suda, Naveen and Civin, Damon and Chandra, Vikas},
  journal={arXiv preprint arXiv:1806.00582},
  year={2018}
}

@article{yang2024regularizing,
  title={Regularizing hidden states enables learning generalizable reward model for llms},
  author={Yang, Rui and Ding, Ruomeng and Lin, Yong and Zhang, Huan and Zhang, Tong},
  journal={Advances in Neural Information Processing Systems},
  volume={37},
  pages={62279--62309},
  year={2024}
}

@inproceedings{jia2024generalizing,
  title={Generalizing reward modeling for out-of-distribution preference learning},
  author={Jia, Chen},
  booktitle={Joint European Conference on Machine Learning and Knowledge Discovery in Databases},
  pages={107--124},
  year={2024},
  organization={Springer}
}

@inproceedings{ye2024openfedllm,
  title={Openfedllm: Training large language models on decentralized private data via federated learning},
  author={Ye, Rui and Wang, Wenhao and Chai, Jingyi and Li, Dihan and Li, Zexi and Xu, Yinda and Du, Yaxin and Wang, Yanfeng and Chen, Siheng},
  booktitle={Proceedings of the 30th ACM SIGKDD conference on knowledge discovery and data mining},
  pages={6137--6147},
  year={2024}
}

@inproceedings{mcmahan2017communication,
  title={Communication-efficient learning of deep networks from decentralized data},
  author={McMahan, Brendan and Moore, Eider and Ramage, Daniel and Hampson, Seth and y Arcas, Blaise Aguera},
  booktitle={Artificial intelligence and statistics},
  pages={1273--1282},
  year={2017},
  organization={PMLR}
}

@inproceedings{huang2021personalized,
  title={Personalized cross-silo federated learning on non-iid data},
  author={Huang, Yutao and Chu, Lingyang and Zhou, Zirui and Wang, Lanjun and Liu, Jiangchuan and Pei, Jian and Zhang, Yong},
  booktitle={Proceedings of the AAAI conference on artificial intelligence},
  volume={35},
  number={9},
  pages={7865--7873},
  year={2021}
}

@inproceedings{zhang2023fedala,
  title={Fedala: Adaptive local aggregation for personalized federated learning},
  author={Zhang, Jianqing and Hua, Yang and Wang, Hao and Song, Tao and Xue, Zhengui and Ma, Ruhui and Guan, Haibing},
  booktitle={Proceedings of the AAAI conference on artificial intelligence},
  volume={37},
  number={9},
  pages={11237--11244},
  year={2023}
}

@article{zhang2020personalized,
  title={Personalized federated learning with first order model optimization},
  author={Zhang, Michael and Sapra, Karan and Fidler, Sanja and Yeung, Serena and Alvarez, Jose M},
  journal={arXiv preprint arXiv:2012.08565},
  year={2020}
}

@inproceedings{collins2021exploiting,
  title={Exploiting shared representations for personalized federated learning},
  author={Collins, Liam and Hassani, Hamed and Mokhtari, Aryan and Shakkottai, Sanjay},
  booktitle={International conference on machine learning},
  pages={2089--2099},
  year={2021},
  organization={PMLR}
}

@article{arivazhagan2019federated,
  title={Federated learning with personalization layers},
  author={Arivazhagan, Manoj Ghuhan and Aggarwal, Vinay and Singh, Aaditya Kumar and Choudhary, Sunav},
  journal={arXiv preprint arXiv:1912.00818},
  year={2019}
}

@article{xu2024dpo,
  title={Is dpo superior to ppo for llm alignment? a comprehensive study},
  author={Xu, Shusheng and Fu, Wei and Gao, Jiaxuan and Ye, Wenjie and Liu, Weilin and Mei, Zhiyu and Wang, Guangju and Yu, Chao and Wu, Yi},
  journal={arXiv preprint arXiv:2404.10719},
  year={2024}
}

@article{tunstall2023zephyr,
  title={Zephyr: Direct distillation of lm alignment},
  author={Tunstall, Lewis and Beeching, Edward and Lambert, Nathan and Rajani, Nazneen and Rasul, Kashif and Belkada, Younes and Huang, Shengyi and Von Werra, Leandro and Fourrier, Cl{\'e}mentine and Habib, Nathan and others},
  journal={arXiv preprint arXiv:2310.16944},
  year={2023}
}

@article{rieke2020future,
  title={The future of digital health with federated learning},
  author={Rieke, Nicola and Hancox, Jonny and Li, Wenqi and Milletari, Fausto and Roth, Holger R and Albarqouni, Shadi and Bakas, Spyridon and Galtier, Mathieu N and Landman, Bennett A and Maier-Hein, Klaus and others},
  journal={NPJ digital medicine},
  volume={3},
  number={1},
  pages={119},
  year={2020},
  publisher={Nature Publishing Group UK London}
}

@article{lin2024limited,
  title={On the limited generalization capability of the implicit reward model induced by direct preference optimization},
  author={Lin, Yong and Seto, Skyler and Ter Hoeve, Maartje and Metcalf, Katherine and Theobald, Barry-John and Wang, Xuan and Zhang, Yizhe and Huang, Chen and Zhang, Tong},
  journal={arXiv preprint arXiv:2409.03650},
  year={2024}
}

@article{li2023policy,
  title={Policy optimization in rlhf: The impact of out-of-preference data},
  author={Li, Ziniu and Xu, Tian and Yu, Yang},
  journal={arXiv preprint arXiv:2312.10584},
  year={2023}
}

@inproceedings{azar2024general,
  title={A general theoretical paradigm to understand learning from human preferences},
  author={Azar, Mohammad Gheshlaghi and Guo, Zhaohan Daniel and Piot, Bilal and Munos, Remi and Rowland, Mark and Valko, Michal and Calandriello, Daniele},
  booktitle={International Conference on Artificial Intelligence and Statistics},
  pages={4447--4455},
  year={2024},
  organization={PMLR}
}

@article{sanh2019distilbert,
  title={DistilBERT, a distilled version of BERT: smaller, faster, cheaper and lighter},
  author={Sanh, Victor and Debut, Lysandre and Chaumond, Julien and Wolf, Thomas},
  journal={arXiv preprint arXiv:1910.01108},
  year={2019}
}

@article{zhang2024tinyllama,
  title={Tinyllama: An open-source small language model},
  author={Zhang, Peiyuan and Zeng, Guangtao and Wang, Tianduo and Lu, Wei},
  journal={arXiv preprint arXiv:2401.02385},
  year={2024}
}

@article{kopf2023openassistant,
  title={Openassistant conversations-democratizing large language model alignment},
  author={K{\"o}pf, Andreas and Kilcher, Yannic and Von R{\"u}tte, Dimitri and Anagnostidis, Sotiris and Tam, Zhi Rui and Stevens, Keith and Barhoum, Abdullah and Nguyen, Duc and Stanley, Oliver and Nagyfi, Rich{\'a}rd and others},
  journal={Advances in neural information processing systems},
  volume={36},
  pages={47669--47681},
  year={2023}
}

\newpage
\appendix
\onecolumn

\section{Notation}
\label{app:notation}

We provide a comprehensive summary of the mathematical notation used throughout this paper in Table~\ref{tab:notation}. This table covers symbols related to federated learning settings, model architectures, hyperparameters, loss functions, and gradient operators.

\begin{table}[h!]
\centering

\caption{Notation Table}
\vspace{-0.5em}
\label{tab:notation}
\footnotesize
\setlength{\tabcolsep}{10pt}
\renewcommand{\arraystretch}{0.98}
\begin{tabular}{p{0.22\textwidth}p{0.5\textwidth}}  
\cmidrule(lr){1-2}  
\textbf{Notation} & \textbf{Description} \\
\cmidrule(lr){1-2}
\multicolumn{2}{l}{\textit{Sets and Data}} \\
$N$ & Number of clients \\
$\mathcal{D}_i$ & Private preference dataset of client $i$ \\
$|\mathcal{D}_i|$ & Size of dataset $\mathcal{D}_i$ \\
$(x, y_w, y_l)$ & Preference triple: prompt, chosen response, rejected response \\
\cmidrule(lr){1-2}
\multicolumn{2}{l}{\textit{Model Parameters}} \\
$\boldsymbol{\Theta}_i$ & Complete model parameters of client $i$ \\
$\mathbf{W}_0 \in \mathbb{R}^{m \times n}$ & Frozen LLM backbone weight matrix \\
$\mathbf{A}_i \in \mathbb{R}^{r \times n}$ & LoRA low-rank matrix A of client $i$ \\
$\mathbf{B}_i \in \mathbb{R}^{m \times r}$ & LoRA low-rank matrix B of client $i$ \\
$\mathbf{M}_i$ & Bottleneck adapter of client $i$ \\
$\mathbf{h}_{1,i}$ & Language modeling head of client $i$ \\
$\mathbf{h}_{2,i}$ & Reward head of client $i$ \\
\cmidrule(lr){1-2}
\multicolumn{2}{l}{\textit{Feature Vectors}} \\
$\mathbf{z}' \in \mathbb{R}^d$ & Hidden state from LoRA-adapted backbone (input to bottleneck) \\
$\mathbf{z} \in \mathbb{R}^d$ & Output from bottleneck adapter (input to both heads) \\
\cmidrule(lr){1-2}
\multicolumn{2}{l}{\textit{Hyperparameters}} \\
$T$ & Total number of communication rounds \\
$t$ & Communication round index \\
$m, n$ & Dimensions of weight matrix $\mathbf{W}_0$ \\
$r$ & Rank of LoRA matrices, $r \ll \min(m, n)$ \\
$d$ & Dimension of hidden states \\
$\eta_h$ & Learning rate for personalized modules \\
$\eta_w$ & Learning rate for LoRA adapters \\
$\beta$ & Temperature parameter in DPO \\
$w_r$ & Reward weight for explicit reward margin \\
$s$ & Adaptive scaling factor \\
$\epsilon$ & Small constant for numerical stability \\
$p_i$ & Aggregation weight, $p_i = \frac{|\mathcal{D}_i|}{\sum_{j=1}^N |\mathcal{D}_j|}$ \\
\cmidrule(lr){1-2}
\multicolumn{2}{l}{\textit{Functions and Operations}} \\
$\mathcal{L}_i(\boldsymbol{\Theta}_i, \mathcal{D}_i)$ & Local loss function of client $i$ \\
$\mathcal{L}_{\text{PDPO}}$ & Personalized DPO loss \\
$\pi_\theta(y|x)$ & Policy distribution under parameters $\theta$ \\
$\pi_{\text{ref}}(y|x)$ & Frozen reference policy \\
$\sigma(\cdot)$ & Sigmoid function \\
$f_{\text{mlp}}(\cdot)$ & Two-layer MLP with Tanh and dropout \\
$f_{\cdot}(\cdot)$ & Bottleneck transformation \\
$\text{Encode}(\cdot)$ & Down-projection in bottleneck \\
$\text{Decode}(\cdot)$ & Up-projection in bottleneck \\
$\text{Pool}(\cdot)$ & Sequence-level pooling \\
$\text{GELU}(\cdot)$ & Gaussian Error Linear Unit \\
$\text{Dropout}(\cdot)$ & Dropout regularization \\
$\text{LayerNorm}(\cdot)$ & Layer normalization \\
\cmidrule(lr){1-2}
\multicolumn{2}{l}{\textit{Preference Margins}} \\
$\Delta_{\text{ir}}$ & Implicit reward margin \\
$\Delta_{\text{er}}$ & Explicit reward margin \\
$\Delta$ & Combined margin, $\Delta = \Delta_{\text{ir}} + w_r \cdot s \cdot \Delta_{\text{er}}$ \\
\cmidrule(lr){1-2}
\multicolumn{2}{l}{\textit{Gradient Operators}} \\
$\nabla_{\mathbf{M}_i}$ & Gradient w.r.t. bottleneck adapter \\
$\nabla_{\mathbf{h}_{1,i}}$ & Gradient w.r.t. LM head \\
$\nabla_{\mathbf{h}_{2,i}}$ & Gradient w.r.t. reward head \\
$\nabla_{(\mathbf{A}_i, \mathbf{B}_i)}$ & Gradient w.r.t. LoRA adapters \\
\cmidrule(lr){1-2}  
\end{tabular}
\end{table}

\clearpage

\section{Preference Learning Data Format: An Illustrative Example}
\label{app:data_format}

In preference-based alignment, training data consists of \textbf{preference triples} $(x, y_w, y_l)$, where:
\begin{itemize}[leftmargin=1.5em]
    \item $x$: the input prompt or context
    \item $y_w$: the \textcolor{blue}{\textbf{chosen}} (preferred) response
    \item $y_l$: the \textcolor{red}{\textbf{rejected}} (dis-preferred) response
\end{itemize}

\noindent
The learning objective is to train a model that assigns higher likelihood or reward to $y_w$ than to $y_l$ given prompt $x$.

\begin{figure}[h!]
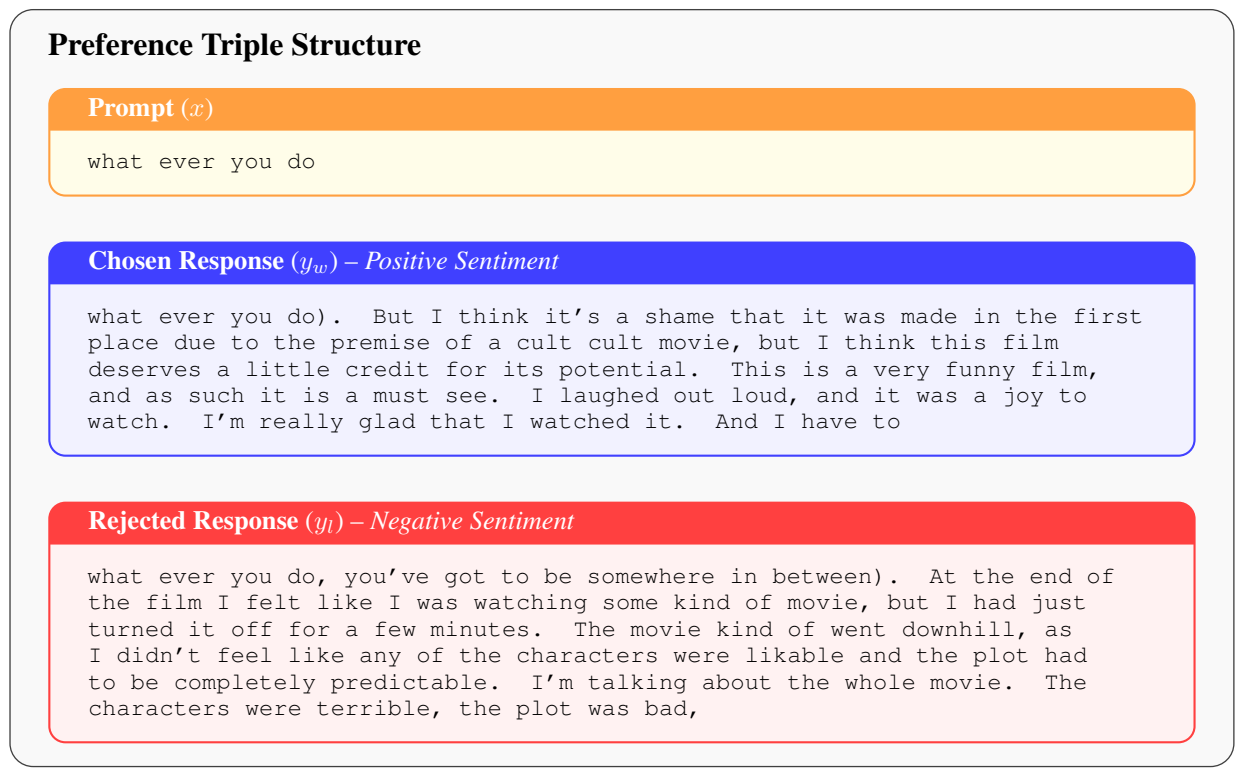

    \centering
    \begin{tcolorbox}[
        colback=gray!5,
        colframe=black!75,
        width=0.95\textwidth,
        arc=3mm,
        boxrule=0.5pt
    ]
    \textbf{\large Preference Triple Structure}
    \vspace{0.3em}
    
    \begin{tcolorbox}[
        colback=yellow!10,
        colframe=orange!75,
        title=\textbf{Prompt} ($x$),
        width=\linewidth,  
        arc=2mm,
        boxrule=0.8pt
    ]
    \small\texttt{what ever you do}
    \end{tcolorbox}
    
    \vspace{0.5em}
    
    \begin{tcolorbox}[
        colback=blue!5,
        colframe=blue!75,
        title=\textbf{Chosen Response} ($y_w$) -- \textit{Positive Sentiment},
        coltitle=white,
        width=\linewidth,  
        arc=2mm,
        boxrule=0.8pt
    ]
    \small\texttt{what ever you do). But I think it's a shame that it was made in the first place due to the premise of a cult cult movie, but I think this film deserves a little credit for its potential. This is a very funny film, and as such it is a must see. I laughed out loud, and it was a joy to watch. I'm really glad that I watched it. And I have to}
    \end{tcolorbox}
    
    \vspace{0.5em}
    
    \begin{tcolorbox}[
        colback=red!5,
        colframe=red!75,
        title=\textbf{Rejected Response} ($y_l$) -- \textit{Negative Sentiment},
        coltitle=white,
        width=\linewidth,  
        arc=2mm,
        boxrule=0.8pt
    ]
    \small\texttt{what ever you do, you've got to be somewhere in between). At the end of the film I felt like I was watching some kind of movie, but I had just turned it off for a few minutes. The movie kind of went downhill, as I didn't feel like any of the characters were likable and the plot had to be completely predictable. I'm talking about the whole movie. The characters were terrible, the plot was bad,}
    \end{tcolorbox}
    
    \end{tcolorbox}
    \caption{An example preference triple from the IMDB dataset. The prompt is a movie review prefix, and the model must learn to prefer the chosen response (positive sentiment, blue) over the rejected response (negative sentiment, red).}
    \label{fig:preference_example}
\end{figure}

To illustrate this data structure, we provide a concrete example from the IMDB sentiment preference dataset in Figure~\ref{fig:preference_example}. The figure shows how a preference triple is organized: the prompt is a movie review prefix, and the two candidate responses represent continuations with opposite sentiments (positive vs. negative).

\clearpage

\section{Proofs for Section~\ref{sec:theory}}
\label{app:theory_proofs}

\subsection{Proof of Theorem 1}
\label{app:theorem1_proof}

\noindent
\textbf{Theorem 1 (Gumbel--Bradley--Terry equivalence).} 
Let $y_w$ and $y_l$ be two responses to a prompt $x$, and let $\mathbf{h}_{2,i}(x,y_w)$ and $\mathbf{h}_{2,i}(x,y_l)$ be their associated estimated rewards from the reward head. 
Finally, let
\[
\begin{aligned}
R_w &\sim \operatorname{Gumbel}\!\left(\mathbf{h}_{2,i}(x,y_w),\,1\right), \\
R_l &\sim \operatorname{Gumbel}\!\left(\mathbf{h}_{2,i}(x,y_l),\,1\right)
\end{aligned}
\]
be independent Gumbel random variables. Then we have
\[
\Pr(R_w - R_l > 0) 
= p_{BT}(y_w \succ y_l \mid x) 
= \sigma\!\big(\Delta_{\text{er}}\big),
\]
where $p_{BT}(y_w \succ y_l \mid x)$ denotes the Bradley–Terry probability and $\Delta_{\text{er}} = \mathbf{h}_{2,i}(x,y_w) - \mathbf{h}_{2,i}(x,y_l)$.

\vspace{0.5em}
\noindent\textbf{Proof.}
Define $I=\arg\max_{j\in\{w,l\}}\{R_j\}$ and abbreviate
$r_w \triangleq \mathbf{h}_{2,i}(x,y_w)$, $r_l \triangleq \mathbf{h}_{2,i}(x,y_l)$.
For a $\operatorname{Gumbel}(\mu,1)$ random variable, the CDF and PDF are
\[
F_\mu(m)=\exp\!\big(-e^{-(m-\mu)}\big), 
\qquad
f_\mu(m)=e^{-(m-\mu)}\,\exp\!\big(-e^{-(m-\mu)}\big).
\]
Then
\[
\Pr(I=w)=\Pr(R_w>R_l)
=\mathbb{E}_{m\sim f_{r_w}}\!\big[\Pr(R_l<m)\big]
=\int_{-\infty}^{+\infty} f_{r_w}(m)\,F_{r_l}(m)\,dm.
\]
Plugging in the expressions of $f_{r_w}$ and $F_{r_l}$ gives
\[
\begin{aligned}
\Pr(I=w)
&=\int_{-\infty}^{+\infty}
e^{-(m-r_w)}\,\exp\!\big(-e^{-(m-r_w)}\big)\,
\exp\!\big(-e^{-(m-r_l)}\big)\,dm \\[2pt]
&=\int_{-\infty}^{+\infty}
e^{r_w-m}\,
\exp\!\Big(-e^{r_w-m}-e^{r_l-m}\Big)\,dm \\[2pt]
&=e^{r_w}\int_{-\infty}^{+\infty}
e^{-m}\,\exp\!\Big(-(e^{r_w}+e^{r_l})e^{-m}\Big)\,dm.
\end{aligned}
\]
Let $t=e^{-m}$ so that $dm=-\,dt/t$, and the limits change from $m\!:\;-\infty\!\to\!+\infty$ to $t\!:\;+\infty\!\to\!0$.
Then
\[
\begin{aligned}
\Pr(I=w)
&=e^{r_w}\int_{+\infty}^{0} t\,\exp\!\big(-(e^{r_w}+e^{r_l})t\big)\,\frac{-\,dt}{t} \\[2pt]
&=e^{r_w}\int_{0}^{+\infty} \exp\!\big(-(e^{r_w}+e^{r_l})t\big)\,dt \\[2pt]
&=\frac{e^{r_w}}{e^{r_w}+e^{r_l}}.
\end{aligned}
\]
This is exactly the Bradley--Terry probability:
\[
\Pr(I=w)=p_{BT}(y_w\succ y_l\mid x)
=\sigma\!\big(r_w-r_l\big)
=\sigma(\Delta_{\text{er}}).
\]
\hfill$\square$
\subsection{Proof of Theorem 2}
\label{app:theorem2_proof}

\noindent
\textbf{Theorem 2 (Reward-correction threshold form).}
Let $y_w$ and $y_l$ be two responses for a prompt $x$, and let $\Delta_{\text{er}} = \mathbf{h}_{2,i}(x,y_w) - \mathbf{h}_{2,i}(x,y_l)$ as above.
For any scalar correction $c \in \mathbb{R}$,
\[
\Pr\!\big(R_w - R_l + c > 0\big) \;=\; \sigma\!\big(c + \Delta_{\text{er}}\big).
\]

\vspace{0.5em}
\noindent\textbf{Proof.}
Let $D \triangleq R_w - R_l \sim \mathrm{Logistic}(\Delta_{\text{er}},1)$ with CDF $F_D(z)=\sigma(z-\Delta_{\text{er}})$. Then
\[
\begin{aligned}
\Pr(D + c > 0)
&= \Pr(D > -c) \\
&= 1 - F_D(-c) \\
&= 1 - \sigma\!\big(-c - \Delta_{\text{er}}\big) \\
&= \tfrac{1}{2} - \tfrac{1}{2}\,\tanh\!\Big(\tfrac{-(c+\Delta_{\text{er}})}{2}\Big) \\
&= \tfrac{1}{2} + \tfrac{1}{2}\,\tanh\!\Big(\tfrac{c+\Delta_{\text{er}}}{2}\Big)
 \;=\; \sigma(c+\Delta_{\text{er}}).
\end{aligned}
\]
\hfill$\square$

\clearpage

\section{Experimental Setup and Implementation Details}
\label{app:exp_setup}

This section provides comprehensive details on the experimental setup, including dataset descriptions and partitioning strategies, model architectures, hyperparameter configurations, and extended experimental results with per-client accuracy breakdowns.

\subsection{Datasets}
\label{app:datasets}

\subsubsection{Dataset Descriptions}

We evaluate \texttt{FedPDPO} under two federated settings: \textbf{intra-domain} and \textbf{cross-domain}.

\paragraph{Intra-Domain Datasets.}
In the intra-domain setting, all clients train on partitions of the same preference dataset:

\textbf{IMDB.} The IMDB sentiment corpus \citep{maas2011learning} consists of English movie reviews with binary sentiment labels, where preferences emphasize sentiment consistency and coherence. We use the complete \texttt{yuasosnin/imdb-dpo} dataset (10k total): \textbf{9{,}000} train, \textbf{1{,}000} test.

\textbf{Code-Vulnerability-Security.} A synthetic dataset targeting secure vs. insecure code across 11 programming languages (C/C++, Python, Java, JavaScript, C\#, PHP, Ruby, Swift, Go, Kotlin, Fortran, Rust, TypeScript), where preferences emphasize code safety and vulnerability mitigation. We use the complete \texttt{CyberNative/Code\_Vulnerability\_Security\_DPO} dataset (4.7k total): \textbf{4{,}194} train, \textbf{466} test.

\paragraph{Cross-Domain Datasets.}
In the cross-domain setting, each client is assigned a distinct dataset from fundamentally different domains:

\textbf{WebGPT.} \citep{nakano2021webgpt} provides long-form question-answering comparisons where preferences emphasize factual accuracy and grounding in web sources. We use a subset from \texttt{openai/webgpt\_comparisons} (19.6k total): \textbf{12{,}000} train, \textbf{1{,}434} test.

\textbf{PyDPO.} A Python code generation dataset where preferences reflect code correctness, efficiency, and best practices, with chosen responses from validated solutions and rejected responses from weaker models. We use \texttt{jondurbin/py-dpo-v0.1} (9.47k total): \textbf{8{,}966} train, \textbf{500} test.

\textbf{UltraFeedback.} \citep{cui2023ultrafeedback} A large-scale instruction-following dataset with multi-dimensional preference annotations covering helpfulness, truthfulness, and instruction adherence. We use \texttt{kaitchup/UltraFeedback-prompt-chosen-rejected} (17.1k total): \textbf{12{,}000} train, \textbf{895} test.

\subsubsection{Non-IID Data Partitioning Strategies}

\begin{figure}[h!]
  \centering
  \includegraphics[width=0.7\columnwidth]{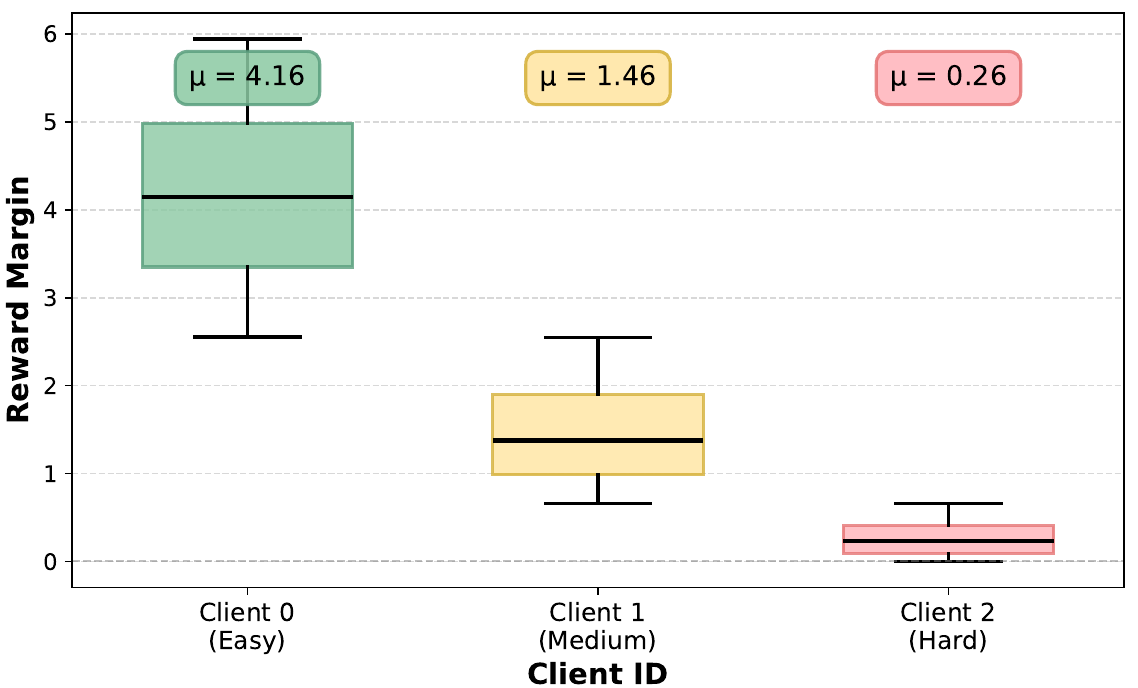}
  \caption{Distribution of reward margins across the IMDB dataset and their non-IID partition into three subsets (easy, medium, hard).}
  \label{fig:reward_margin_distribution}
\end{figure}

To simulate realistic statistical heterogeneity, we design dataset-specific non-IID partition strategies.

\paragraph{Reward-Margin Based Partition (IMDB).}
We create non-IID splits by leveraging the reward margin: $\text{margin} = r_{\text{chosen}} - r_{\text{rejected}}$. Samples are sorted by margin in descending order and partitioned into contiguous shards. For 3 clients, we assign easy (avg.\ margin $\approx 4.16$), medium ($\approx 1.46$), and hard ($\approx 0.26$) subsets. For 10 clients, the sorted list is evenly divided into ten shards from easiest to hardest. Figure~\ref{fig:reward_margin_distribution} illustrates the reward margin distribution and partition strategy.

\paragraph{Language-Based Partition (Code-Vulnerability-Security).}
We partition clients by programming language. For 3 clients: Client0 (Web \& Scripting: JavaScript, PHP, Python, Ruby, TypeScript), Client1 (Enterprise \& Mobile: Java, C\#, Kotlin, Swift), Client2 (Systems \& HPC: C/C++, Go, Fortran, Rust). For 11 clients, each receives one language with 90\%/10\% train/test split.

\paragraph{Cross-Domain Assignment.}
Each client is assigned a distinct dataset: Client0 (WebGPT), Client1 (PyDPO), Client2 (UltraFeedback).

\subsection{Model Architectures}
\label{app:models}

We employ two backbone models to validate generalizability across different scales. All personalized components use configurable hidden dimensions to balance expressiveness and efficiency.

\subsubsection{Backbone Models}

\paragraph{DistilGPT2.}
We adopt DistilGPT2 \citep{sanh2019distilbert} as the primary backbone. This distilled GPT-2 variant consists of 6 transformer layers with hidden dimension $d=768$. The backbone remains frozen, with only LoRA adapters applied for parameter-efficient fine-tuning.

\paragraph{TinyLlama-1.1B.}
We additionally evaluate TinyLlama-1.1B \citep{zhang2024tinyllama}, a 1.1B parameter model with 22 layers and hidden dimension 2048. Results are in Table~\ref{tab:heterogeneous_tinyllama}.

\subsubsection{LoRA Adapters}

Following LoRA \citep{hu2022lora}, we inject trainable low-rank matrices into the frozen backbone. For weight matrix $\mathbf{W}_0 \in \mathbb{R}^{m \times n}$:
\[
\mathbf{W} = \mathbf{W}_0 + \mathbf{B}\mathbf{A},
\]
where $\mathbf{A} \in \mathbb{R}^{r \times n}$ and $\mathbf{B} \in \mathbb{R}^{m \times r}$ with rank $r=8$, $\alpha=16$, dropout $=0.1$. LoRA adapters target \texttt{c\_attn}, \texttt{c\_proj}, \texttt{c\_fc} modules and are the only shared components aggregated across clients.

\subsubsection{Bottleneck Adapter}

The bottleneck adapter $\mathbf{M}_i$ fuses global knowledge from LoRA with local representations. Given hidden state $\mathbf{z}' \in \mathbb{R}^{d}$ from the LoRA-adapted backbone (where $d=768$ for DistilGPT2):
\[
\mathbf{z} = \text{LayerNorm}\big(\mathbf{z}' + f_{\cdot}(\mathbf{z}')\big),
\]
where $f_{\cdot}(\mathbf{z}') = \text{Decode}\!\left(\text{Dropout}\!\left(\text{GELU}\!\left(\text{Encode}(\mathbf{z}')\right)\right)\right)$.

The projections are $\text{Encode}: \mathbb{R}^{768} \to \mathbb{R}^{256}$ (down-projection) and $\text{Decode}: \mathbb{R}^{256} \to \mathbb{R}^{768}$ (up-projection). This architecture compresses the 768-dimensional features into a 256-dimensional latent space before projecting back to the original dimension, allowing selective feature modulation. We apply GELU activation between projections and dropout $=0.1$ for regularization. The residual connection ensures refinement rather than replacement of the original features. Each client maintains a private bottleneck adapter that is not shared during aggregation.

\subsubsection{Personalized Dual Heads}

Each client maintains two personalized prediction heads receiving features from the bottleneck adapter:

\paragraph{Language Modeling Head ($\mathbf{h}_{1,i}$).}
A standard linear projection from the bottleneck output $\mathbf{z} \in \mathbb{R}^{768}$ to vocabulary size, producing token-level logits:
\[
\text{logits} = \mathbf{h}_{1,i}(\mathbf{z}) \in \mathbb{R}^{50257},
\]
where the vocabulary size $|\mathcal{V}|=50257$ for GPT-2. These logits define the policy distribution $\pi_\theta(y|x)$ used in DPO's implicit reward formulation.

\paragraph{Reward Head ($\mathbf{h}_{2,i}$).}
Estimates scalar preference scores for input-output pairs. Given a sequence of hidden states $\{\mathbf{z}_1, \ldots, \mathbf{z}_T\}$ from the bottleneck adapter, we first apply mean pooling:
\[
\bar{\mathbf{z}} = \frac{1}{T}\sum_{t=1}^{T} \mathbf{z}_t \in \mathbb{R}^{768}.
\]
The pooled representation is then passed through a two-layer MLP:
\[
\mathbf{h}_{2,i}(x,y) = \mathbf{W}_2 \cdot \tanh\!\left(\text{Dropout}(\mathbf{W}_1 \cdot \bar{\mathbf{z}} + \mathbf{b}_1)\right) + b_2,
\]
where $\mathbf{W}_1 \in \mathbb{R}^{256 \times 768}$, $\mathbf{W}_2 \in \mathbb{R}^{1 \times 256}$, $\mathbf{b}_1 \in \mathbb{R}^{256}$, $b_2 \in \mathbb{R}$, and dropout $=0.1$. The network transforms the 768-dimensional pooled features through a 256-dimensional hidden layer to produce a scalar reward score. This explicit reward complements the implicit reward from log-probabilities in the PDPO objective (Eq.~\eqref{eq:pdpo_loss}).

Both heads are fully personalized and remain local to each client, enabling adaptation to client-specific preference distributions without compromising privacy.

\subsubsection{Reward Model for PPO Baselines}

For PPO baselines, we use OpenAssistant's pre-trained reward model \citep{kopf2023openassistant}: \texttt{OpenAssistant/reward-model-deberta-v3-large-v2}. This DeBERTa-v3-large model predicts human preference rankings, trained on WebGPT comparisons, feedback summaries, synthetic instruction pairs, and Anthropic's HH-RLHF dataset. The reward model remains frozen and globally shared across all clients during PPO training. In contrast, \texttt{FedPDPO} trains its reward head jointly with the policy in a self-contained manner.

\subsubsection{Model Configuration Summary}
 \vspace{0.5em}
\begin{table}[h!]
\centering
\caption{Model architecture configurations for \texttt{FedPDPO}.}
\label{tab:model_config}
\small
\begin{tabular}{lcc}
\toprule
\textbf{Component} & \textbf{DistilGPT2} & \textbf{TinyLlama-1.1B} \\
\midrule
Backbone dimension ($d$) & 768 & 2048 \\
Number of layers & 6 & 22 \\
Backbone parameters & 82M & 1.1B \\
Frozen during training & \checkmark & \checkmark \\
\midrule
LoRA rank ($r$) & 8 & 8 \\
LoRA $\alpha$ & 16 & 16 \\
LoRA dropout & 0.1 & 0.1 \\
LoRA trainable params & $\sim$0.3M & $\sim$0.8M \\
\midrule
Bottleneck hidden dim & 256 & 256 \\
Bottleneck dropout & 0.1 & 0.1 \\
Bottleneck trainable params & $\sim$0.4M & $\sim$1.0M \\
\midrule
LM head trainable params & $\sim$38M & $\sim$102M \\
Reward head hidden dim & 256 & 256 \\
Reward head trainable params & $\sim$0.2M & $\sim$0.5M \\
\midrule
\textbf{Total trainable (per client)} & $\sim$39M & $\sim$104M \\
\textbf{Aggregated (LoRA only)} & $\sim$0.3M & $\sim$0.8M \\
\bottomrule
\end{tabular}
\end{table}

Table~\ref{tab:model_config} summarizes the architectural configurations.

The architecture aggregates only LoRA parameters ($<$1\% of trainable parameters) while maintaining rich personalization through local modules. Configurable hidden dimensions enable flexible capacity-efficiency trade-offs.

\subsection{Hyperparameter Configurations}
\label{app:hyperparameters}

We report the key hyperparameters for our federated training setups. Unless otherwise specified, all methods adopt AdamW \citep{loshchilov2017decoupled} as the optimizer with $(\beta_1,\beta_2)=(0.9,0.999)$, weight decay $=0.01$, gradient clipping at $1.0$, and a batch size of $8$ with gradient accumulation of $2$ steps. The backbone is \texttt{Distilgpt2} with LoRA adapters of rank $r=8$, bottleneck adapters with hidden dimension $256$ and dropout $=0.1$, and a reward head with hidden dimension $256$ (two-layer MLP). For baselines (\texttt{FedAvg}, \texttt{Per-FedAvg}, \texttt{FedAMP}, \texttt{FedPer}, \texttt{FedRep}), all hyperparameters are aligned with \texttt{FedPDPO} to ensure fairness, except for the method-specific parameters noted below.

\subsubsection{Intra-Domain Setting}

\paragraph{Federated configuration.}
\begin{itemize}[leftmargin=1.5em]
    \item \textbf{Number of clients:} 3 or 10 for IMDB, 3 or 11 for Code-Vulnerability-Security
    \item \textbf{Communication rounds:} 20
    \item \textbf{Client participation ratio:} 1.0
    \item \textbf{Local training epochs:} 1 for LoRA modules; \texttt{FedPDPO} additionally trains personalized modules (bottleneck adapter and dual heads) for 1 epoch
    \item \textbf{Learning rates:} Grid search over $\{1\!\times\!10^{-4}, 5\!\times\!10^{-5}, 3\!\times\!10^{-5}, 8\!\times\!10^{-5}, 1\!\times\!10^{-3}\}$; selected $1\!\times\!10^{-4}$ for personalized modules and $5\!\times\!10^{-5}$ for LoRA modules
    \item \textbf{LR decay:} factor $0.8$, minimum LR $=10^{-6}$; decay triggered when DPO loss increases
\end{itemize}

\paragraph{DPO-specific parameters.}
\begin{itemize}[leftmargin=1.5em]
    \item \emph{Common:} $\beta=0.1$, reference-free $=$ False
    \item \emph{\texttt{FedPDPO}-specific:} reward weight ratio $w_r$ schedule (linear from $0.5$ to $1.5$, baseline $0.8$), adaptive reward scaling (EMA momentum $0.95$), sample exclusion threshold ($k=2$, maximum exclusion ratio $0.5$)
\end{itemize}
\paragraph{PPO-specific parameters.}
For computational efficiency in federated settings, PPO baselines adopt a simplified offline variant that operates on pre-collected preference pairs, using the frozen reward model (deberta-v3-large-v2) to score responses and computing advantages from reward differences. All PPO baselines use the same LoRA architecture as \texttt{FedPDPO} for fair comparison.
\begin{itemize}[leftmargin=1.5em]
    \item KL coefficient: $0.1$
    \item Value function coefficient: $0.5$
    \item Clipping parameter: $\epsilon=0.2$
    \item Entropy coefficient: $c_e=0.01$
    \item GAE parameter: $\lambda=0.95$
    \item PPO epochs per round: $4$; Mini-batch size: $4$; Target KL: $0.01$
\end{itemize}

\paragraph{LoRA configuration.}
\begin{itemize}[leftmargin=1.5em]
    \item Rank $r=8$, $\alpha=16$, dropout $=0.1$
    \item Target modules: \texttt{c\_attn}, \texttt{c\_proj}, \texttt{c\_fc}
\end{itemize}

\paragraph{Baseline-specific parameters.}
\begin{itemize}[leftmargin=1.5em]
    \item \textbf{FedAvg:} standard averaging, no additional parameters
    \item \textbf{Per-FedAvg:} inner-loop LR $0.01$, outer-loop LR $0.001$, inner-loop steps $=5$
    \item \textbf{FedAMP:} similarity temperature $\tau=0.5$, attention weight floor $0.05$
    \item \textbf{FedPer:} first $2$ transformer layers shared globally, remaining layers and head personalized
    \item \textbf{FedRep:} backbone LR $5\!\times\!10^{-5}$, head LR $1\!\times\!10^{-4}$
\end{itemize}

\subsubsection{Cross-Domain Setting}

\paragraph{Federated configuration.}
\begin{itemize}[leftmargin=1.5em]
    \item \textbf{Number of clients:} 3 (WebGPT, PyDPO, UltraFeedback)
    \item \textbf{Communication rounds:} 10
    \item \textbf{Client participation ratio:} 1.0
    \item \textbf{Local training epochs:} 3 for LoRA modules; \texttt{FedPDPO} additionally trains personalized modules for 1 epoch
    \item \textbf{Learning rates:} Same grid search protocol as intra-domain setting
    \item \textbf{LR decay:} factor $0.8$, minimum LR $=10^{-6}$
\end{itemize}

\paragraph{DPO-specific, PPO-specific, LoRA, and Baseline-specific parameters.}
Same as in the intra-domain setting.

\subsection{Extended Experimental Results}
\label{app:extended_results}

\subsubsection{Impact of Key Hyperparameters}

To systematically validate our architectural and hyperparameter choices for \texttt{FedPDPO}, we conduct comprehensive sensitivity analysis across five benchmark settings: cross-domain average (Cross), IMDB with 3 and 10 clients, and Code-Vulnerability-Security with 3 and 11 clients. For each hyperparameter, we vary one factor while keeping others fixed at default values (LoRA rank=8, reward head dim=256, adapter dim=256, $\beta$=0.1) and all other training configurations consistent with Appendix~\ref{app:hyperparameters}, reporting the best test accuracy achieved across all communication rounds.

\begin{figure}[h!]
  \centering
   \includegraphics[width=0.6\columnwidth]{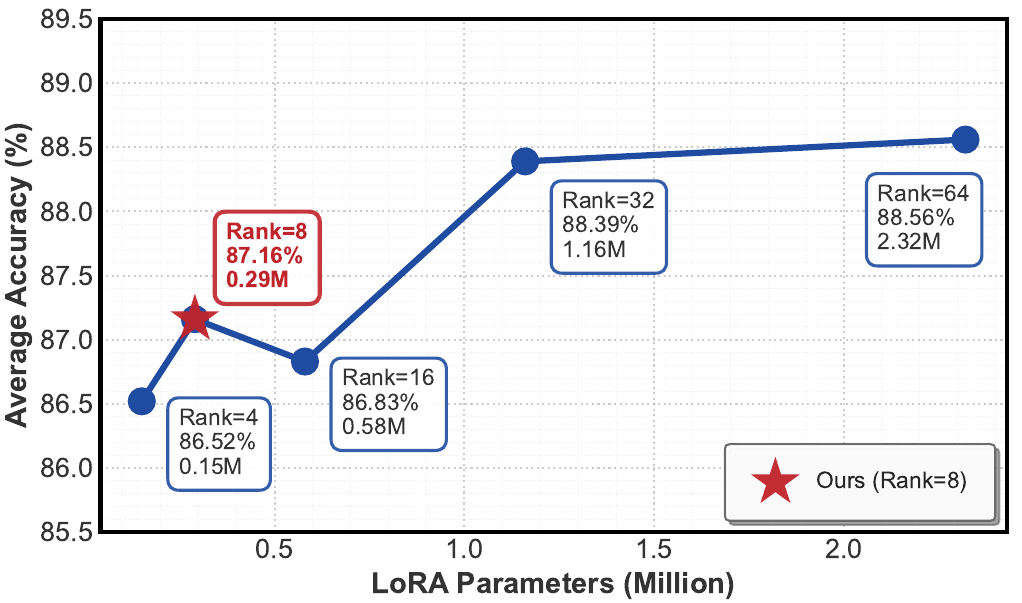}
   \vspace{-0.5em}
  \caption{Average accuracy (\%) varies with LoRA rank across five benchmark settings.}
  \label{fig:ablation_lora}
  \vspace{-0.8em}
\end{figure}

\paragraph{LoRA Rank.}
Table~\ref{tab:ablation_lora} shows the impact of LoRA rank $r \in \{4, 8, 16, 32, 64\}$ on model performance. Rank=8 excels on cross-domain (73.44\%), IMDB settings (87.41\%, 85.23\%), and Code-3 (96.92\%), achieving strong average performance (87.16\%). Notably, Rank=16 underperforms Rank=8 despite having twice the parameters (86.83\% vs 87.16\%), suggesting diminishing returns with moderate rank increases. While Rank=64 achieves the highest average (88.56\%) and Rank=32 also shows gains (88.39\%), higher ranks require 4$\times$ and 8$\times$ more communication overhead respectively. Rank=8 provides the optimal efficiency-performance trade-off for federated scenarios where communication is the bottleneck. Figure~\ref{fig:ablation_lora} visualizes the accuracy-parameter trade-off across different LoRA ranks.

\begin{table}[h!]
\centering
\small
\caption{Test accuracy (\%) under varying LoRA ranks across five  benchmark settings.}
\vspace{-0.5em}
\label{tab:ablation_lora}
\setlength{\tabcolsep}{5pt}
\renewcommand{\arraystretch}{1.05}
\begin{tabular}{ccccccc}
\toprule
\textbf{Rank} & \textbf{IMDB-3} & \textbf{IMDB-10} & \textbf{Code-3} & \textbf{Code-11} & \textbf{Cross} & \textbf{Avg} \\
\midrule
4  & 86.24 & 83.69 & \underline{96.98} & 92.99 & 72.70 & 86.52 \\
8  & 87.41 & 85.23 & 96.92 & 92.80 & 73.44 & 87.16 \\
16 & \textbf{87.92} & 84.19 & 96.58 & 94.12 & 73.36 & 86.83 \\
32 & \underline{87.68} & \textbf{87.54} & \textbf{97.24} & \underline{95.26} & \underline{74.23} & \underline{88.39} \\
64 & 87.66 & \underline{87.44} & \underline{96.98} & \textbf{96.21} & \textbf{74.51} & \textbf{88.56} \\
\bottomrule
\end{tabular}
\vspace{-1em}
\end{table}

\begin{figure}[h!]
\vspace{-0.5em}
  \centering
  \includegraphics[width=0.6\columnwidth]{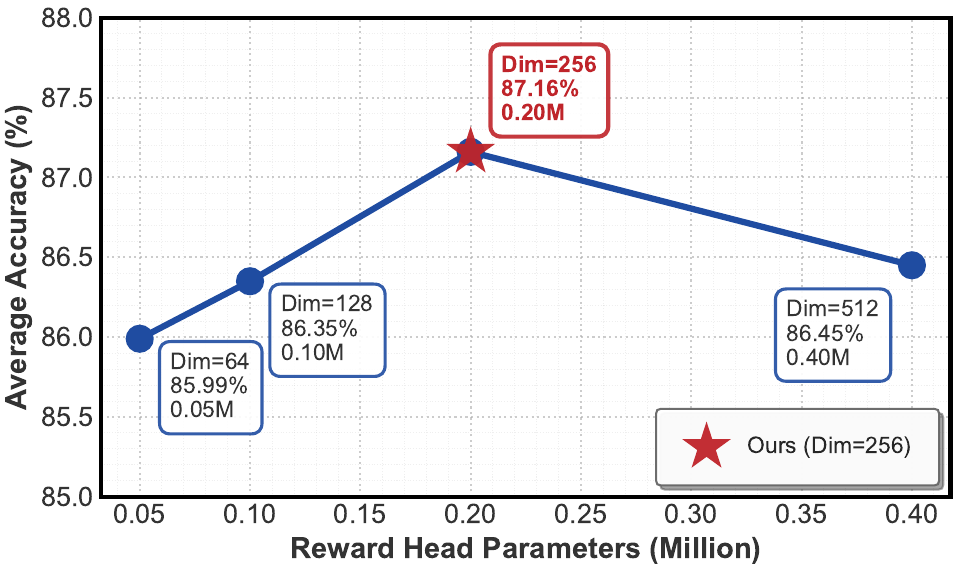}
  \vspace{-0.5em}
  \caption{Average accuracy (\%) varies with reward head dimension across five benchmark settings.}
  \label{fig:ablation_reward}
  \vspace{-1.5em}
\end{figure}

\paragraph{Reward Head Dimension.}
Table~\ref{tab:ablation_reward} analyzes the impact of reward head capacity with hidden dimensions $\{64, 128, 256, 512\}$. Dim=256 achieves the best average performance (87.16\%) and shows strongest results on cross-domain (73.44\%), both IMDB settings (87.41\%, 85.23\%), and Code-3 (96.92\%). Smaller dimensions (64, 128) cause noticeable degradation especially on cross-domain ($-1.04\%$, $-1.37\%$) and IMDB-10 ($-3.27\%$, $-2.64\%$), while Dim=512 shows diminishing returns likely due to overfitting, demonstrating that Dim=256 provides optimal capacity for explicit reward modeling without overfitting. Figure~\ref{fig:ablation_reward} illustrates the performance-capacity relationship of the reward head.

\begin{table}[h!]
\centering
\small
\vspace{0em}
\caption{Test accuracy (\%) under varying reward head dimensions across five benchmark settings.}
\vspace{-0.5em}
\label{tab:ablation_reward}
\setlength{\tabcolsep}{5pt}
\renewcommand{\arraystretch}{1.05}
\begin{tabular}{ccccccc}
\toprule
\textbf{Dim} & \textbf{IMDB-3} & \textbf{IMDB-10} & \textbf{Code-3} & \textbf{Code-11} & \textbf{Cross} & \textbf{Avg} \\
\midrule
64  & 86.56 & 81.96 & 96.23 & 92.80 & \underline{72.40} & 85.99 \\
128 & \underline{87.16} & 82.59 & \underline{96.81} & \textbf{93.13} & 72.07 & 86.35 \\
256 & \textbf{87.41} & \textbf{85.23} & \textbf{96.92} & 92.80 & \textbf{73.44} & \textbf{87.16} \\
512 & 87.08 & \underline{84.01} & 96.36 & \underline{92.97} & 71.83 & \underline{86.45} \\
\bottomrule
\end{tabular}
\vspace{-0.7em}
\end{table}

\begin{figure}[h!]
  \centering
  \includegraphics[width=0.6\columnwidth]{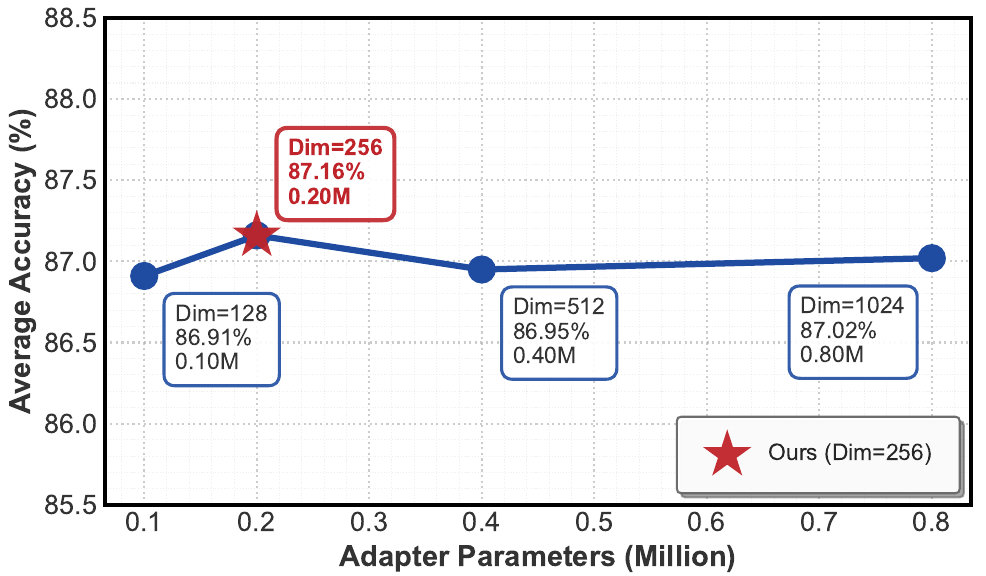}
  \vspace{-0.5em}
  \caption{Average accuracy (\%) varies with bottleneck adapter dimension across five benchmark settings.}
  \label{fig:ablation_adapter}
  \vspace{-1.5em}
\end{figure}

\paragraph{Bottleneck Adapter Dimension.}
Table~\ref{tab:ablation_adapter} examines the bottleneck adapter's compression ratio with dimensions $\{128, 256, 512, 1024\}$. Dim=256 achieves the best average performance (87.16\%) and excels on cross-domain (73.44\%), IMDB-3 (87.41\%), and Code-3 (96.92\%), implementing an effective 3:1 compression ratio (768$\rightarrow$256$\rightarrow$768) that balances global LoRA features with local backbone representations. Too small bottleneck (Dim=128) causes information loss, while too large bottleneck (Dim=512, 1024) reduces regularization effect and shows inconsistent performance across datasets. Figure~\ref{fig:ablation_adapter} shows the impact of compression ratio on feature fusion quality.

\begin{table}[h!]
\centering
\small
\caption{Test accuracy (\%) under varying bottleneck adapter dimensions across five benchmark settings.}
\vspace{-0.5em}
\label{tab:ablation_adapter}
\setlength{\tabcolsep}{5pt}
\renewcommand{\arraystretch}{1.05}
\begin{tabular}{ccccccc}
\toprule
\textbf{Dim} & \textbf{IMDB-3} & \textbf{IMDB-10} & \textbf{Code-3} & \textbf{Code-11} & \textbf{Cross} & \textbf{Avg} \\
\midrule
128 & 86.64 & \underline{86.23} & 96.03 & 93.35 & 72.11 & 86.91 \\
256 & \textbf{87.41} & 85.23 & \textbf{96.92} & 92.80 & \textbf{73.44} & \textbf{87.16} \\
512 & \underline{87.07} & 85.02 & \underline{96.52} & \underline{93.91} & \underline{72.24} & 86.95 \\
1024 & 86.48 & \textbf{86.43} & 96.35 & \textbf{94.08} & 71.76 & \underline{87.02} \\
\bottomrule
\end{tabular}
\vspace{-0.5em}
\end{table}

\begin{figure}[h!]
 \vspace{-0.5em}
  \centering
  \includegraphics[width=0.6\columnwidth]{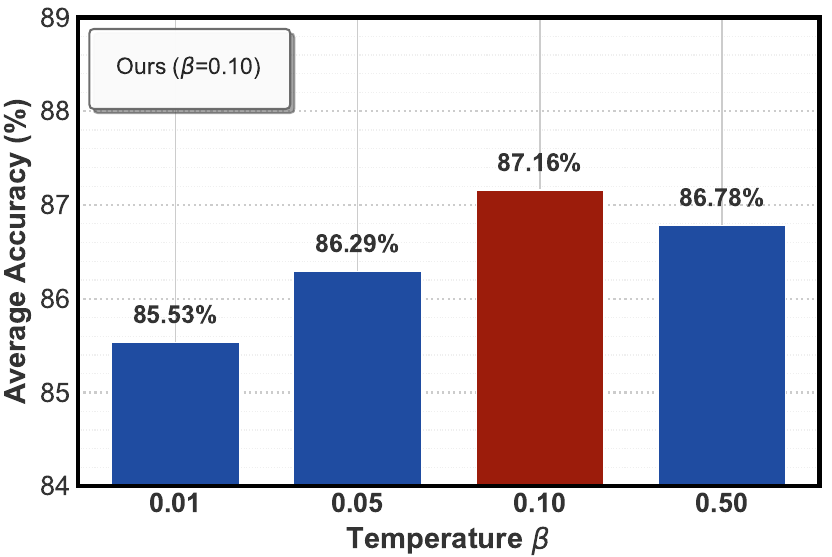}
  \vspace{-0.5em}
  \caption{Average accuracy (\%) varies with DPO temperature $\beta$ across five benchmark settings.}
  \label{fig:ablation_beta}

\end{figure}

\paragraph{DPO Temperature $\beta$.}
Table~\ref{tab:ablation_beta} studies the sensitivity to DPO temperature $\beta \in \{0.01, 0.05, 0.10, 0.50\}$. $\beta$=0.10 achieves the best average performance (87.16\%) and excels on cross-domain (73.44\%), both IMDB settings (87.41\%, 85.23\%), and Code-3 (96.92\%). Too small $\beta$ (0.01) causes severe underfitting ($-1.93\%$ average), especially on IMDB datasets ($-1.76\%$ on IMDB-3, $-2.68\%$ on IMDB-10), while too large $\beta$ (0.50) leads to overfitting on training preferences, hurting cross-domain generalization ($-1.68\%$) and IMDB-10 ($-1.77\%$), though showing gains on Code datasets, demonstrating that $\beta$=0.10 provides optimal balance between preference exploitation and KL regularization. Figure~\ref{fig:ablation_beta} visualizes the impact of temperature on DPO training stability.

\begin{table}[h!]
\vspace{-0.5em}
\centering
\small
\caption{Test accuracy (\%) under varying DPO temperature $\beta$ across five benchmark settings.}
\vspace{-0.5em}
\label{tab:ablation_beta}
\setlength{\tabcolsep}{5pt}
\renewcommand{\arraystretch}{1.05}
\begin{tabular}{ccccccc}
\toprule
\textbf{$\beta$} & \textbf{IMDB-3} & \textbf{IMDB-10} & \textbf{Code-3} & \textbf{Code-11} & \textbf{Cross} & \textbf{Avg} \\
\midrule
0.01 & 85.65 & 82.55 & 95.94 & 91.85 & 71.67 & 85.53 \\
0.05 & 86.84 & \underline{83.96} & 96.37 & 92.23 & \underline{72.04} & 86.29 \\
0.10 & \textbf{87.41} & \textbf{85.23} & \underline{96.92} & \underline{92.80} & \textbf{73.44} & \textbf{87.16} \\
0.50 & \underline{87.25} & 83.46 & \textbf{97.33} & \textbf{94.12} & 71.76 & \underline{86.78} \\
\bottomrule
\end{tabular}
\vspace{-1em}
\end{table}

\paragraph{Summary.}
Our sensitivity analysis validates the following optimal configuration for \texttt{FedPDPO}: LoRA rank=8, reward head dim=256, adapter dim=256, and $\beta$=0.10. This configuration achieves strong performance across diverse federated scenarios while prioritizing cross-domain robustness and communication efficiency—the core objectives of federated learning—over marginal in-domain accuracy gains that come at significant computational cost.

\subsubsection{Per-Client Accuracy Breakdown}

Table~\ref{tab:homogeneous_results} provides the detailed per-client accuracy breakdown for the intra-domain experiments with 3 clients reported in Table~\ref{tab:compare-results} of the main paper. These results demonstrate that \texttt{FedPDPO} consistently achieves the highest accuracy across all clients under both reward-margin–based partitioning (IMDB) and language-based partitioning (Code-Vulnerability-Security), validating the effectiveness of our personalized dual-head design and bottleneck adapter fusion mechanism under heterogeneous data distributions.

\begin{table}[ht!]
  \centering
  \vspace{-0.5em}
  \caption{Per-client test accuracy (\%) under intra-domain settings with 3 clients. Left: Reward-Margin Partition (IMDB). Right: Language-Based Partition (Code-Vulnerability-Security). Results are mean $\pm$ std over 5 runs.}
  \label{tab:homogeneous_results}
  \setlength{\tabcolsep}{4pt}
  \renewcommand{\arraystretch}{1.08}
  \begin{tabular}{@{\hspace{2pt}}l@{\hspace{8pt}}cccc@{\hspace{10pt}}cccc@{\hspace{2pt}}}
    \toprule
    \multirow{2}{*}{\textbf{Method}}
    & \multicolumn{4}{c}{\textbf{IMDB}} 
    & \multicolumn{4}{c}{\textbf{Code-Vulnerability-Security}} \\
    \cmidrule(lr){2-5} \cmidrule(lr){6-9}
    & \textbf{Client0} & \textbf{Client1} & \textbf{Client2} & \textbf{Avg} 
    & \textbf{Client0} & \textbf{Client1} & \textbf{Client2} & \textbf{Avg} \\
    \midrule
    FedAvg$_{\text{(+PPO)}}$ & 93.87$_{\pm0.20}$ & 80.94$_{\pm0.24}$ & 67.28$_{\pm0.27}$ & 80.70
    & 96.43$_{\pm0.17}$ & 91.26$_{\pm0.20}$ & 86.52$_{\pm0.24}$ & 91.40 \\
    FedAvg$_{\text{(+DPO)}}$ & 95.54$_{\pm0.18}$ & 83.19$_{\pm0.22}$ & 69.47$_{\pm0.25}$ & 82.73
    & 98.21$_{\pm0.14}$ & 93.45$_{\pm0.17}$ & 88.80$_{\pm0.21}$ & 93.49 \\
    \midrule
    Per-FedAvg$_{\text{(+PPO)}}$ & 94.56$_{\pm0.24}$ & 81.08$_{\pm0.28}$ & 68.45$_{\pm0.33}$ & 81.36
    & 95.84$_{\pm0.21}$ & 92.37$_{\pm0.24}$ & 88.12$_{\pm0.31}$ & 92.11 \\
    Per-FedAvg$_{\text{(+DPO)}}$ & 96.13$_{\pm0.21}$ & \underline{83.33}$_{\pm0.25}$ & \underline{70.73}$_{\pm0.30}$ & \underline{83.40}
    & 97.62$_{\pm0.18}$ & 94.64$_{\pm0.21}$ & 90.36$_{\pm0.27}$ & 94.21 \\
    \midrule
    FedAMP$_{\text{(+PPO)}}$ & 93.21$_{\pm0.23}$ & 80.12$_{\pm0.30}$ & 67.89$_{\pm0.36}$ & 80.41
    & 95.67$_{\pm0.22}$ & 90.58$_{\pm0.26}$ & 87.34$_{\pm0.33}$ & 91.20 \\
    FedAMP$_{\text{(+DPO)}}$ & 95.54$_{\pm0.20}$ & 82.56$_{\pm0.27}$ & 70.14$_{\pm0.33}$ & 82.75 
    & 97.45$_{\pm0.19}$ & 92.86$_{\pm0.23}$ & 89.84$_{\pm0.30}$ & 93.38 \\
    \midrule
    FedPer$_{\text{(+PPO)}}$ & \underline{96.78}$_{\pm0.23}$ & 78.34$_{\pm0.32}$ & 66.93$_{\pm0.39}$ & 80.68 
    & 96.89$_{\pm0.18}$ & 93.12$_{\pm0.23}$ & 90.54$_{\pm0.31}$ & 93.52 \\
    FedPer$_{\text{(+DPO)}}$ & 95.24$_{\pm0.24}$ & 80.60$_{\pm0.29}$ & 69.15$_{\pm0.36}$ & 81.66 
    & \underline{98.21}$_{\pm0.16}$ & \underline{95.24}$_{\pm0.20}$ & 92.19$_{\pm0.28}$ & \underline{95.21} \\
    \midrule
    FedRep$_{\text{(+PPO)}}$ & 94.38$_{\pm0.25}$ & 79.12$_{\pm0.31}$ & 66.78$_{\pm0.35}$ & 80.09
    & 96.32$_{\pm0.18}$ & 91.84$_{\pm0.25}$ & 90.43$_{\pm0.28}$ & 92.86 \\
    FedRep$_{\text{(+DPO)}}$ & 96.13$_{\pm0.22}$ & 81.37$_{\pm0.28}$ & 69.05$_{\pm0.32}$ & 82.18 
    & \underline{98.21}$_{\pm0.15}$ & 94.05$_{\pm0.22}$ & \underline{92.71}$_{\pm0.25}$ & 95.00 \\
    \midrule
    \rowcolor{gray!10}
    \textbf{\texttt{FedPDPO}}
    & \textbf{98.21}$_{\pm0.16}$ & \textbf{89.23}$_{\pm0.20}$ & \textbf{74.80}$_{\pm0.19}$ & \textbf{87.41} 
    & \textbf{98.81}$_{\pm0.12}$ & \textbf{98.21}$_{\pm0.15}$ & \textbf{93.75}$_{\pm0.18}$ & \textbf{96.92} \\
    \bottomrule
  \end{tabular}
\end{table}

\end{document}